\documentclass{article}
\usepackage{arxiv}

\usepackage[utf8]{inputenc}
\usepackage[T1]{fontenc}
\usepackage{amsmath,amsfonts,amssymb}
\usepackage{algorithmic}
\usepackage[linesnumbered,longend,ruled,lined,boxed,commentsnumbered]{algorithm2e}
\usepackage{array}
\usepackage[caption=false,font=normalsize,labelfont=sf,textfont=sf]{subfig}
\usepackage{textcomp}
\usepackage{stfloats}
\usepackage{url}
\usepackage{verbatim}
\usepackage{graphicx}
\usepackage{xcolor}
\usepackage{float}
\usepackage{multirow}
\usepackage{hhline}
\usepackage{tabularx}
\usepackage{threeparttable}
\usepackage{booktabs}
\usepackage{makecell}
\usepackage{cite}
\usepackage[colorlinks,bookmarksopen,bookmarksnumbered,citecolor=blue,linkcolor=blue,urlcolor=blue]{hyperref}
\usepackage{cleveref}
\usepackage{soul}
\setstcolor{red}
\graphicspath{{./}}

\title{Hierarchical Reinforcement Learning in StarCraft Micromanagement with Influence Maps and Cluster-based Scripts}

\author{
\normalfont
\begin{tabular}{c}
Chunhui Bai$^{1,3,4}$, Changhe Li$^{2,5,*}$, Dequan Li$^{5}$, Xinye Cai$^{2,5}$, Shengxiang Yang$^{6}$\\
\texttt{baichunhui@cug.edu.cn}, \texttt{changhe.lw@gmail.com}, \texttt{dqli@aust.edu.cn}\\
\texttt{2024084@aust.edu.cn}, \texttt{syang@dmu.ac.uk}
\end{tabular}\\[0.4em]
\begin{minipage}{0.96\textwidth}
\footnotesize
$^{1}$ School of Artificial Intelligence and Automation, China University of Geosciences, Wuhan 430074, China\\
$^{2}$ State Key Laboratory of Digital Intelligent Technology for Unmanned Coal Mining, Anhui University of Science \& Technology, Huainan 232001, China\\
$^{3}$ Hubei Key Laboratory of Advanced Control and Intelligent Automation for Complex Systems, Wuhan 430074, China\\
$^{4}$ Engineering Research Center of Intelligent Technology for Geo-Exploration, Ministry of Education, Wuhan 430074, China\\
$^{5}$ School of Artificial Intelligence, Anhui University of Science \& Technology, Hefei 231131, China\\
$^{6}$ Computer Science and Informatics, De Montfort University, Leicester, U.K.\\
$^{*}$ Corresponding author.
\end{minipage}
}

\begin{document}
\maketitle

\begin{abstract}

Real-time strategy (RTS) games present significant AI challenges, characterized by expansive state-action spaces arising from multi-unit coordination in continuous battlefields, and sparse delayed rewards stemming from final win/lose signals. Existing approaches face a trade-off between managing the dimensionality explosion of joint actions and maintaining the interpretability of complex state representations. This complexity is further intensified by the limitation of traditional hierarchical structures in adaptively decomposing tasks into effective tactical modules. Such difficulties are compounded by the black-box nature of deep learning models and their reliance on sparse rewards, which together result in limited sample efficiency and a lack of decision-making transparency. To address these limitations, this paper proposes HRL-IM/CBS, a hierarchical reinforcement learning framework with influence map hashing and cluster-based scripts for StarCraft micromanagement. Influence map hashing encodes global battlefield situations into compact hexadecimal codes, capturing spatial control and relative advantage. Cluster-based scripts enable dynamic local coordination through adaptive unit partitioning. The hierarchical multi-Q-table architecture decomposes decision-making into upper-level clustering strategy selection and lower-level tactical execution, with reward allocation providing dense learning signals. Experiments across six asymmetric scenarios demonstrate competitive performance against deep RL baselines while offering advantages in sample efficiency and interpretability through transparent Q-table representations.
\end{abstract}



\section{Introduction}
Real-time strategy (RTS) games represent one of the most challenging domains for artificial intelligence research, combining real-time large-scale decision-making and long-term strategic planning requirements \cite{yannakakis_artificial_2018}. Popular titles such as StarCraft II have established themselves as standard benchmarks for AI research, featuring multi-agent coordination, strong adversarial interactions and hierarchical tactical management that mirror many real-world problems \cite{ontanon_survey_2013}.

One of the challenges in RTS AI lies in achieving efficient learning under the circumstances of large state–action space and sparse delayed rewards \cite{Ontanon2024}. Specifically, the enormous state-action space stems from continuous two-dimensional battlefields and multi-unit parameterized actions, resulting in extremely low efficiency of random sampling. Long-term planning tasks in RTS games yield only win/lose signals at episode termination, creating sparse and delayed rewards, making it difficult to accurately assess intermediate action values and allocate rewards. Due to the aforementioned difficulties, the simulation-based training cost will become extremely expensive, necessitating a balance between comprehensive exploration and sample efficiency.

Existing approaches address the challenge of efficient learning from different perspectives. From the perspective of mitigating large-scale challenges, heuristic state-action space processing \cite{critch_combining_2020, marino_Evolving_2019} effectively reduces the scale of problems by sacrificing small accuracy. Hierarchical reinforcement learning \cite{zhou_hierarchical_2021, jiang2025hierarchical} decomposes complex problems to handle large-scale decisions, but it faces difficulties in high storage costs and adaptive hierarchy discovery. From the perspective of balancing computational resource and sample efficiency, deep learning methods represented by neural networks and attention mechanisms \cite{xie_semicentralized_2022, ye2025three} demonstrate powerful performance through model expression capacity and parallel computing ability, but suffer from limitations including high computational overhead and insufficient interpretability. Sample-efficient multi-agent reinforcement learning \cite{hu2023rethinking, manandhar2024reinforcement, qi2026distributed} improves data utilization efficiency from an algorithmic level, but struggles to achieve a balance between low-cost deployment and wide adaptability.

To address the existing limitations, this paper proposes a hierarchical reinforcement learning with influence maps and cluster-based scripts (HRL-IM/CBS) for RTS micromanagement. The contributions of HRL-IM/CBS are as follows:

\begin{itemize}
    \item Proposed influence map hashing and cluster-based scripts can effectively mitigate the impact of state and action space explosion. Influence map hashing encodes unit distribution as a matrix to capture global features and replaces raw influence image storage with histogram equalization to reduce enormous memory and retrieval costs. Cluster-based scripts execute within locally clustered combat subspaces, balancing action space dimensionality with localized tactical coordination.
    \item A hierarchical multi-Q-table framework combined with a reward allocation mechanism is adopted to decompose the flat long-term planning task into hierarchical levels, while mitigating the impact of sparse delayed rewards and enhancing the multi-local coordination. Ablation studies and comparisons across asymmetric experimental scenarios demonstrate clear advantages in training efficiency.
    \item Granularity-adjustable combat space clustering demonstrates interpretability by identifying locally dense clusters and latent cooperative relationships. By integrating fragmented local knowledge into a unified representation through combat space clustering, multi-perspective effectiveness and mechanism analysis confirm that the proposed HRL-IM/CBS framework affords a novel and interpretive analysis of tactical preferences.
\end{itemize}

The remainder of this paper is organized as follows. Section~\ref{Related work} provides an overview of the previous related work. Section~\ref{Methodologies} details the proposed HRL-IM/CBS framework, including influence map hashing, cluster-based script design, and the hierarchical multi-table Q-learning architecture. Section~\ref{Experimental Setup and Design} describes the experimental protocol, scenario design, and parameter settings. Section~\ref{Experimental Results and Analysis} presents the experimental results of the ablation study, quantitative comparisons, as well as effectiveness and mechanism analysis. Section~\ref{Conclusion} concludes the paper and outlines future research directions.

\section{Related Work}
\label{Related work}

In this section, a brief review is provided of how existing StarCraft AI approaches tackle the challenges posed by the vast state–action space and the inefficiency of exploration under sparse and delayed rewards.

\subsection{Mitigating Large-Scale Difficulties}
\subsubsection{State-action Abstraction}
The vast state-action space poses a severe challenge to the difficulty of solving decision-making problems, especially in search algorithms. In previous research on RTS games, especially in micromanagement, influence maps and scripts are two typical methods to reduce state-action space \cite{dockhorn2023state}, demonstrating the advantages of fast computation and strong interpretability.

For state space processing, influence maps provide a simplified representation of the game state in RTS games. It facilitates spatial reasoning by identifying formation boundaries and choke points \cite{uriarte_Kiting_2012}, as well as by capturing tactical spatial features such as the military strength of teams \cite{micic_Developing_2011} and the dangerousness of surrounding locations \cite{shantia_Connectionist_2011}. Influence maps can be widely integrated with various algorithms to enhance pathfinding \cite{park_mcts_2015} and decision-making \cite{lu_intelligent_2022}, leveraging their advantages in reducing memory usage and enhancing computational efficiency. For StarCraft, influence maps have been used to capture spatial information for outcome prediction \cite{sanchez-ruiz_machine_2017}, or guide unit movement efficiently by combining with Bayesian Models \cite{synnaeve_Bayesian_2011c} and heuristic search \cite{critch_combining_2020}. However, most influence map-based methods either directly support decision-making or integrate feature extraction and few studies have attempted to perform learning in an abstract state space to accelerate training.

For action space processing, to efficiently reduce the search space of tree-based search methods, scripts that hard-code human experience were proposed and proven highly effective \cite{yang_experimental_2022}. Similar approaches were later referred to as macro actions \cite{xu_Macro_2019, liu_Efficient_2022a} or action abstractions \cite{moraes_Action_2018, marino_Evolving_2019, moraes_asymmetric_2018} in subsequent research.

Portfolio Greedy Search (PGS) \cite{churchill_portfolio_2013} utilizes greedy mountain climbing with fixed scripts (NOK-AV and Kiter), outperforming Alpha-Beta \cite{churchill_fast_2012} and UCT \cite{balla_UCT_2009} search in certain situations. To address tactical inflexibility, group-based scripts \cite{justesen_script-_2014} demonstrates superior efficiency compared to individual assignment and better corresponds to human gameplay patterns. Following this group-based approach, Stratified Strategy Selection (SSS) \cite{lelis_stratified_2017} reduces search costs by combining unit type features with abstract actions. To alleviate local optimal limitations in PGS and SSS, An asymmetric abstract action search algorithm \cite{moraes_asymmetric_2018} has been proposed to implement restricted unit allocation. However, these script assignment strategies generally rely solely on unit types or statistical characteristics, neglecting dynamic local coordination.

\subsubsection{Hierarchical Reinforcement Learning}
Hierarchical reinforcement learning has been flexibly applied to various large-scale decision-making problems, including cooperative lane change \cite{liang2022hierarchical}, multi-objective energy management \cite{xu2022hierarchical} and routing algorithms \cite{yang2024hierarchical}. The nature of multi-agent tasks with multi-level complexity in RTS games makes hierarchical and modular architectures highly applicable. Additionally, the hierarchical approach alleviates the exponential growth of joint action space in single-layer models as the number of agents increases, thereby improving exploration in complex environments.

Various hierarchical approaches have been developed to tackle specific challenges in RTS games, demonstrating effectiveness in capturing tactical knowledge \cite{zhou_hierarchical_2021}, balancing local-global cooperation \cite{xie_semicentralized_2022} and introducing macro-action abstraction for lightweight computing \cite{jiang2025hierarchical}. However, these hierarchical architectures face fundamental limitations in automatic hierarchy discovery and heavy reliance on human expert knowledge, resulting in insufficient adaptability in dynamic environments. Adaptive hierarchical decomposition will play an important role in dynamic local collaboration in RTS games.

\subsection{Balancing Computational Resource and Sample Efficiency}
RL techniques have been widely applied in StarCraft. While the research community is developing high-performance RTS AI with DRL, MARL and XRL, it is also striving to utilize sample-efficient RL to achieve high training effectiveness at a lower training cost.

\subsubsection{Deep Reinforcement Learning}
With the development of computing power, DRL has gained advantages in handling complex high-dimensional tasks \cite{barros2024deep}. DeepMind's AlphaStar \cite{vinyals_grandmaster_2019} has achieved grandmaster-level performance in StarCraft II using actor-critic methods with LSTM structures to present human-like strategies. However, despite TStarbot-X \cite{han_tstarbot-x_2021} attempts to reduce computing resources, the high computing cost still poses a challenge. To address these limitations, lightweight neural networks like Light-Q-Network (LQN) and Binary-Q-Network (BQN) for resource-constrained training \cite{li_Accelerating_2020} have been proposed. Additionally, 3D attention mechanisms have been utilized to integrate multi-dimensional information for accurate tactical assessment in complex RTS scenarios \cite{ye2025three}.

Most DRL approaches mentioned above require not only extensive iterations of learning from scratch, but the form of a high dimensional data tensor even makes it hard to interpret the performance of the algorithm. Especially at the tactical and micromanagement levels, model effectiveness requires methods for interpreting learned knowledge explicitly.

\subsubsection{Multi-Agent and Explainable Reinforcement Learning}

Multi-agent reinforcement learning has significantly advanced RTS game AI \cite{wang2025intelligent}. Traditional methods like BiCNet \cite{peng_multiagent_2017} and COMA \cite{foerster_counterfactual_2018} establish foundational unit collaboration frameworks but show limited generalization. Parameter sharing approaches like PS-MAGDS \cite{shao_starcraft_2019} improve generalization through parameter sharing and curriculum learning yet suffer from unit avoidance strategies. More recent hierarchical methods like MAHGAC \cite{li2025multi} employ graph attention mechanisms to capture complex cooperative-competitive relationships, but face challenges in training efficiency and computational scalability in complex RTS environments.

Despite advances in MARL for RTS games, fundamental limitations persist in computational scalability, multi-level strategic coordination and training efficiency. Furthermore, A thorough understanding of the behavior and mechanism of learned agents remains a huge challenge in most existing multi-agent reinforcement learning methods. 

The opacity of DRL and MARL has sparked interest in explainable reinforcement learning (XRL). In RTS XRL, intrinsic interpretability methods incorporate comprehensible data representations into algorithm design, such as reward decomposition \cite{panda_Explainable_2025} and derivation tree \cite{marino_Evolving_2022}. In contrast, post-hoc analysis approaches provide explanatory summaries for the decision behaviors of trained black-box RTS models, such as semantic clustering-based counterfactual demonstration methods \cite{mathes_CODEX_2023}. However, these approaches rely on simple scripts or natural language within basic scenarios, lacking multi-level explanations for complex interaction scenarios.

\subsubsection{Sample-efficient Reinforcement Learning}
In recent years, the research community has focused on sample-efficient reinforcement learning in RTS games to achieve high training efficiency at a low training cost. Invalid action masking \cite{hou2023exploring} and monotonicity constraint \cite{hu2023rethinking} prevent invalid exploration by restricting the policy space, but these approaches may limit adaptability to dynamic environmental changes and reduce the discoverability of potentially optimal policies. Executing rehearsal with opponent behavior prediction \cite{manandhar2024reinforcement} filters out high-value samples, but the performance of this method depends on the accuracy of the prediction function and the rehearsal cost will increase the impact of large-scale actions.

Various sparse reward handling techniques have been developed for RTS games, including curriculum-based reward shaping \cite{endla2025deep}, intrinsic curiosity combined with Option-Based Hierarchical PPO (OB-HPPO) \cite{jiang_OBHPPO_2024} and multi-dimensional intrinsic reward \cite{qi2026distributed}. While reward shaping uses manually designed rewards and intrinsic motivation employs agent-generated rewards, both approaches face limitations. Reward shaping may produce unnatural behaviors and poor dynamic adaptation, while intrinsic motivation methods suffer from limited transparency in reward interpretation and unpredictable exploration patterns.

\begin{figure*}[t]
	\centering
	\includegraphics[width=\linewidth]{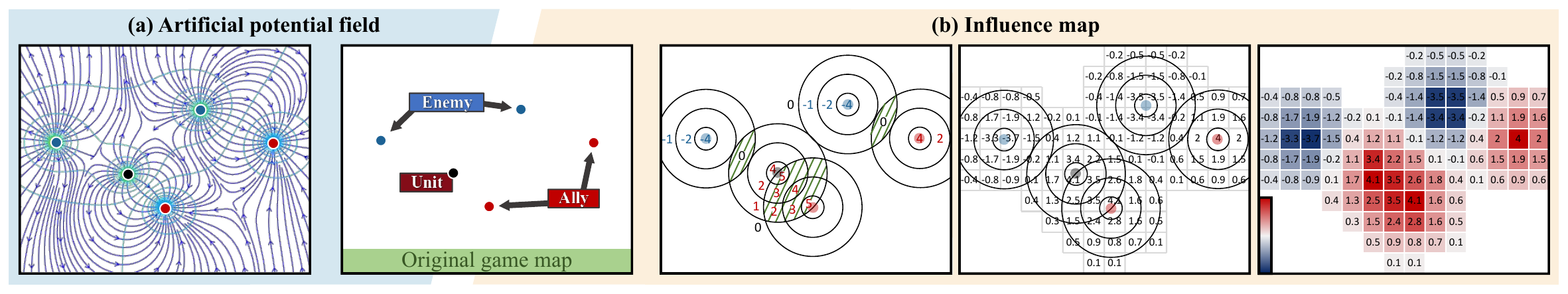}
    \vspace{-2em}
    \caption{Illustration of APF and influence map in RTS games. The black point represents the unit. Red points represent allies, while blue ones represent enemies. (a) For APF, the direction of the arrow represents the navigation to the unit. (b) For influence map, the sphere of influence generated by each unit, where self units generate positive values and enemies generate negative ones, is discretized on a grid map to create the overall influence map.}
    \label{fig:APF-IM}
\end{figure*}

Existing methods attempt to address large-scale challenges and achieve sample efficiency, but they still have limitations. State–action abstraction offers efficient compression but struggles with unit-distribution encoding and script adaptability. Hierarchical architectures effectively decompose strategic tasks but face challenges in automatic hierarchy discovery and human expertise dependency, limiting spatial collaboration. DRL provides computational advantages but suffers from high retraining costs and poor interpretability. Sample-efficient methods focus on resource-constrained training but lack complex scenario adaptability. Consequently, current approaches still struggle to balance lightweight computation, efficient training and strong interpretability.

\section{Methodologies}
\label{Methodologies}
This section presents the HRL-IM/CBS framework for RTS micromanagement. Influence map hashing and cluster-based scripts reduce state-action space complexity, while a hierarchical multi-table Q-learning architecture with local reward allocation enables efficient and interpretable learning.

\subsection{Design Rationale}

The HRL-IM/CBS framework is grounded in a principled hierarchical decomposition that separates global clustering decisions from local tactical execution. The design rationale for the state and action abstractions, and the hierarchical learning architecture is presented as follows.

\subsubsection{State Abstraction via Influence Map Hashing}
State abstraction aims to reduce the cardinality of the state space while preserving information essential for optimal decision-making \cite{li_Unified_2006}. According to the $Q^*$-irrelevance abstraction criterion, states that share the same optimal action-value functions should be mapped to the same abstract representation.

In RTS micromanagement, this equivalence is often manifested as spatial symmetries. Optimal tactical decisions remain invariant under global translation or rotation of combat units. To computationally approximate this property, an abstraction function $\phi: \mathcal{O} \rightarrow \mathcal{S}_{hash}$ is designed based on influence map hashing.

Unlike raw observations, influence map-based RGB histograms function as a proxy for the relative spatial density of forces. This representation is inherently insensitive to absolute coordinates. For any two game observations $O_1$ and $O_2$, the influence map hashing ensures that $\phi(O_1) = \phi(O_2)$ if the normalized tactical distributions represented by histograms $\hat{h}_1$ and $\hat{h}_2$ satisfy:
\begin{equation}
	d_{edit}(\hat{h}_1, \hat{h}_2) < \epsilon
\end{equation}
where $d_{edit}$ denotes the edit distance between discretized histogram bins, and $\hat{h}$ is obtained through 8-bin equalization as detailed in Section~\ref{sec:State space based on Influence Map hashing}. The resulting $\mathcal{S}_{hash}$ has significantly reduced cardinality compared to $\mathcal{O}$, facilitating efficient Q-learning.

\subsubsection{Action Abstraction via Cluster-based Scripts}
Action abstraction is essential for managing the exponential complexity of joint action spaces in RTS micromanagement. A script-based approach, pioneered and validated in RTS combat research \cite{churchill_fast_2012}, provides the lower-level action space for cluster-level tactical execution.

The script set $\Omega$ comprises pre-defined tactical abstractions, each encapsulating domain expertise for maneuvers such as focus-fire, kiting, and retreat. Each cluster $i$ independently selects a script from the shared set:
\begin{equation}
    \omega_i \in \Omega, \quad i = 1, \ldots, C_k
\end{equation}
where $C_k$ denotes the number of clusters. Each cluster maintains a local Q-table $\mathcal{Q}^{l}_i$ to learn optimal script selection. This decentralized design ensures high reactivity while decoupling decision complexity from the total number of units.

\subsubsection{Hierarchical Q-Learning Architecture}
The architecture employs a hierarchical value decomposition with separate learning at global and local scales. Let $\mathcal{Q}^{u}(s, C_s)$ denote the upper-level value function, where $s \in \mathcal{S}_{hash}$ represents the global state encoded by influence map hashing and $C_s$ denotes the clustering strength. Let $\mathcal{Q}^{l}_i(s_i, \omega)$ denote the lower-level value function for cluster $i$, where $s_i$ represents the local state derived from intra-cluster features and $\omega \in \Omega$ represents the selected script. The hierarchical update rules are formulated as:
\begin{equation}
    \label{eq:upper_q}
    \begin{aligned}
        \mathcal{Q}^{u}(s, C_s) \leftarrow & \mathcal{Q}^{u}(s, C_s) + \alpha \Big[ r_g \\
        & + \gamma \max_{C'_s} \mathcal{Q}^{u}(s', C'_s) - \mathcal{Q}^{u}(s, C_s) \Big]
    \end{aligned}
\end{equation}
\begin{equation}
    \label{eq:lower_q}
    \begin{aligned}
        \mathcal{Q}^{l}_i(s_i, \omega) \leftarrow & \mathcal{Q}^{l}_i(s_i, \omega) + \alpha \Big[ r_{l,i} \\
        & + \gamma \max_{\omega'} \mathcal{Q}^{l}_i(s'_i, \omega') - \mathcal{Q}^{l}_i(s_i, \omega) \Big]
    \end{aligned}
\end{equation}
where $r_g$ and $r_{l,i}$ represent global and local rewards, respectively. This decomposition enables modular learning while maintaining tactical coordination.

\subsection{State Space Based on Influence Map Hashing}
\label{sec:State space based on Influence Map hashing}
\subsubsection{Artificial Potential Field and Influence Map}
In fields of robot path planning, Artificial Potential Field (APF), inspired by the interaction forces between charged particles, navigates robots by constructing virtual potential fields, which has advantages of simple design, real-time responsiveness, and high computational efficiency. Similarly, RTS games involve a large number of path planning tasks, including micromanagement. As shown in Figure~\ref{fig:APF-IM}(a), target positions and obstacles can be represented as attractive and repulsive points respectively. Units determine their direction of movement based on the gradient of the potential field, ultimately approaching the target position and avoiding obstacles.

However, in RTS games, path planning tasks are often complicated scenarios where there may be multiple or no exact target locations. The single path provided by the APF may not be able to adapt to dynamic scenarios or meet actual needs. Therefore, the influence map has made improvements in this regard, which considers units as influence points that generate ripples of influence around them, divides the map into a grid map and calculates the influence value for each grid, and ultimately obtains an influence matrix, as illustrated in Figure~\ref{fig:APF-IM}(b).

Square influence maps are used to simplify calculations and highlight enemy units' distribution impact. Self units have positive influence values ([16,9,4,1]), while enemy units have negative ones ([-16,-9,-4,-1]). Each unit's influence range is a 7×7 square area, with absolute influence values decreasing from 16 at the center to 1 at the edges.

\subsubsection{Influence Map Hashing}
Rather than constructing a high-dimensional tensor with features such as the combat space, the quantity, attribute and absolute coordinates of units, influence map abstracts the combat situation into a low-dimensional space. Compared to the decentralized or partially centralized local observation space of multi-agent systems, influence map focuses on the global features of the map, which helps enhance the comprehensiveness and accuracy of decision-making.

\begin{figure}[htbp]
    \centering
    \includegraphics[width=0.88\linewidth]{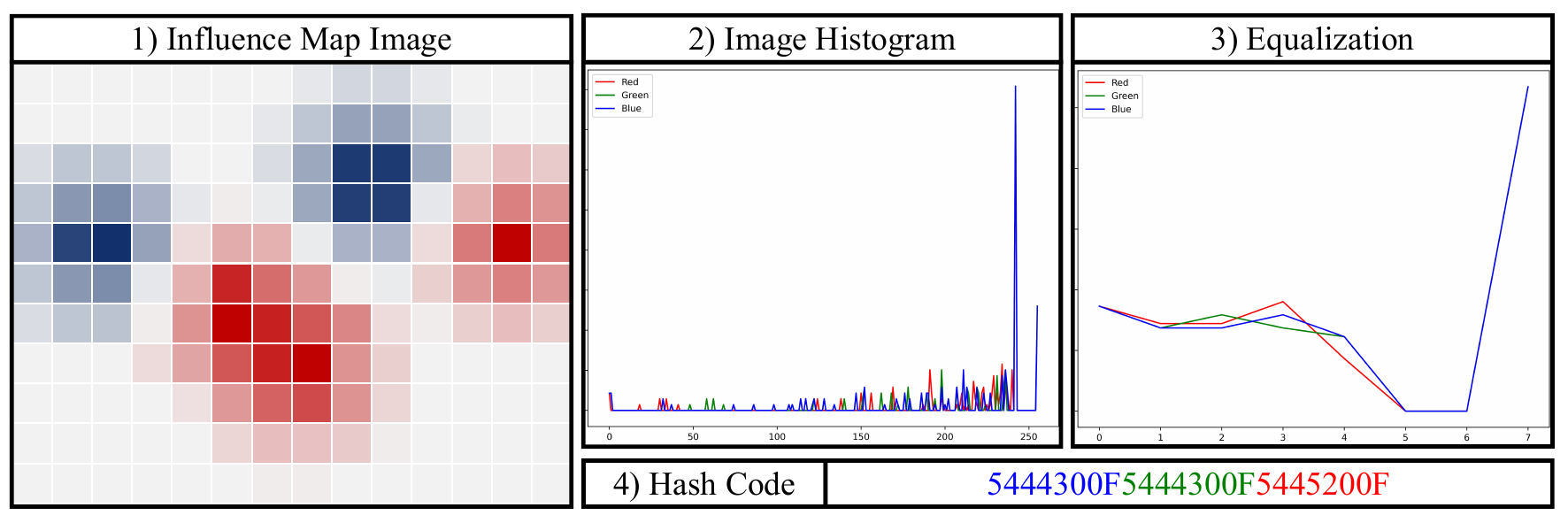}
    \caption{Illustration of influence map hashing. The distribution of units on the original game map is expressed into an influence map image. Through 8 bins histogram equalization, the image is represented as a reduced-length hash code, so that only a 3×8-bit hash code is needed to represent a state.}
    \label{fig:IM_hash}
\end{figure}

However, the distinguishability of states in RTS games should be ensured that each state should be unique and distinct in the whole state space, while minor errors caused by grid influence map should categorize identical states together to ensure the effectiveness of training. In RTS games, there often exist situations where the absolute positions of units change while the relative distribution between opposing sides remains the same, which involves the task of identifying similarities in influence map. In addition, it is necessary to encode the influence map data uniquely to avoid the significant memory and retrieval costs associated with storing all influence map images when dividing similar states. Based on this, this paper utilizes RGB histogram features that are insensitive to image rotation and translation to normalize the influence map, as illustrated in Figure~\ref{fig:IM_hash}.

Take Figure~\ref{fig:IM_hash} as an example, the main steps include: 1) Expressing the distribution of units on the original game map to an influence map value matrix and an influence map image. 2) Extracting the RGB components of the influence map image and calculating the RGB histogram. 3) Performing histogram equalization so that the RGB channels values will be normalized from 256 bins to 8 bins. 4) Finally, the normalized RGB values will be hashed as 3×8-bit hexadecimal hash code of equal length, and be updated into the original state space. So far, the image data has been replaced with 24-bit encoding that can be retrieved by hash. States sampled through interaction with the environment will be added into the state space, which can be referred to as exploration of the state space. The pseudocode for updating states obtained through exploration into the original state space is shown in Algorithm \ref{alg:addstate}. It should be noted that the $d_{edit}(\cdot,\cdot)$ function is implemented by instantiating the edit (Levenshtein) distance, which serves to quantify state similarity and to determine whether a given state has already been sampled. This metric is sufficient to identify similar states and yields a compact and diverse set of representative states.

\renewcommand{\algorithmcfname}{Algorithm}
\begin{algorithm}
    \caption{Update State Space}\label{alg:addstate}
    \renewcommand{\thealgocf}{}
    \SetAlgoLined
    \KwIn{  Maximum number of game step $K$}
    \KwIn{  Current game step $k$}
    \KwIn{  Original state space $\mathcal{S}_{k-1}=\{s_1, s_2, ..., s_{n}\}$ }
    \KwIn{  Game observation information $O_{k}$}
    \KwIn{  Distance threshold $\epsilon$ }
    \KwOut{  Explored state space $\mathcal{S}_{k}=\{s_1, s_2, ..., s_{n}, s_{k}\}$ or $\{s_1, s_2, ..., s_{n}\}$}
    
    \If{$k > 1$}
    {
        
        $s_{k} \gets$ encode\rule{0.15cm}{0.4pt}IM\rule{0.15cm}{0.4pt}matrix$(O_{k})$

        $is\_new \gets$ True

        \ForEach {$s_{i}$ in $\mathcal{S}_{k-1}$}
        {
            
            	\If{$d_{edit}(s_{i},s_{k}) \leq \epsilon$}
                {
				    $is\_new \gets$ False
				    \textbf{break}
                }
            
        }

            \If{$is\_new$}
            {
                $\mathcal{S}_{k} \gets \mathcal{S}_{k-1} \cup \{s_{k}\}$
            }
        
    }

    \SetKwFunction{MyFunction}{encode\_IM\_matrix}
        \SetKwProg{Fn}{Function}{:}{}

        \Fn{\MyFunction{$O_{k}$}} {
            \tcp{Encode an influence map matrix into hash code}

            $m_{k} \gets $calculate$\_$IM$\_$matrix$(O_{k})$

            $g_{k} \gets $Generate RGB image $g_{k}$ based on $m_{k}$

            $h_{k} \gets$ Calculate histogram of the image $h_{k}$ based on $g_{k}$
    
            $\hat{h_{k}} \gets h_{k}.histogram$\_$equalization()$
    
            $s_{k} \gets \hat{h_{k}}.hash$\_$code()$
            
            \Return{$s_{k}$}
        }

\end{algorithm}

In summary, the multi-dimensional combat situation is encoded into a compact set of hash codes that depend solely on the relative positions of units. This representation simultaneously distinguishes distinct states, clusters similar ones, and yields a substantial reduction in data volume. It should be noted that the resulting histogram will be affected by the color scheme. To ensure reproducibility, a single, fixed color scheme is employed throughout the entire training pipeline.

\subsection{Script-based Action Space and Combat Space Clustering}
\label{sec:Action space based on scripts and combat space clustering}
\subsubsection{Script}
In RTS games micro-management, it is often necessary to execute a series of precise actions on multiple units, which results in a multidimensional and large-scale action space. In order to reduce the size of the action space and alleviate the problem of large action space brought by micro-actions on multi-unit combat, as well as enhance the tactical rationality of units, this paper adopts scripts to form the action space.

The so-called scripts are not precise actions that specifically target the micro-level actions of individual units, but pre-coded tactical strategies that deploy actions for multiple units simultaneously. The agent selects combinations of scripts for units within a certain segment of frames to accomplish various combat missions.

\subsubsection{Combat Space Clustering}
\begin{figure}[htbp]
    \centering
    \includegraphics[width=0.88\linewidth]{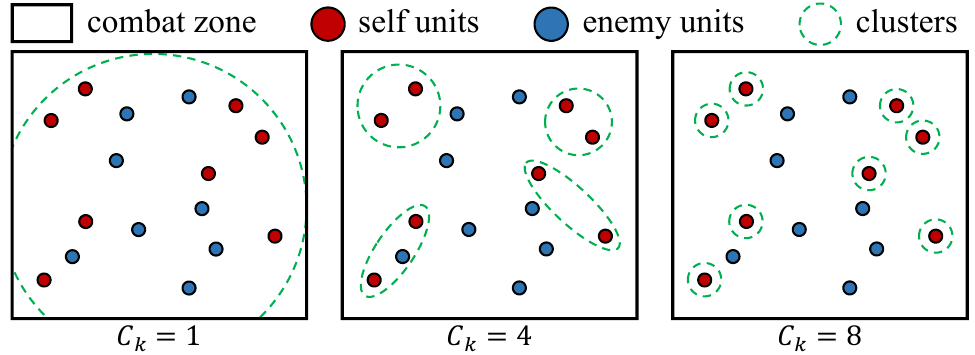}
    \caption{Illustration of combat space clustering under different $C_k$. The red point represents self unit. The blue point represents enemy unit. The green dashed circle represent clusters, where units execute the same script.}
    \label{fig:cluster}
\end{figure}

The combat space refers to the instantaneous relative spatial distribution of all combat units, which may contain locally dense groups or potential cooperative relationships. When facing complex battle scenarios, units often need to adopt different tactics to demonstrate cooperation and synergy, thus maximizing team rewards. To ensure that different units can take different actions or scripts at a certain moment, making local collaboration more effective and reasonable, it is necessary to cluster the combat space. One feasible method is to cluster the combat space based on the relative distance between units. This paper adopts k-means clustering to implement the combat space clustering, and the clustering results under different numbers of clusters, denoted as $C_k$, are illustrated in Figure~\ref{fig:cluster}.

Units in one cluster execute the same scripts, while different clusters may adopt different tactics. Choosing a reasonable $C_k$ is crucial to balance local collaboration and global cooperation. A lower $C_k$ reduces the number of clusters and strengthens unit consistency within clusters, while a higher $C_k$ allows for more tactic diversity. Specifically, $C_k=1$ signifies unified combat space clustering, where all units use the same script. In contrast, $C_k=8$ (the total number of self units) enables each unit to independently select scripts.

In addition, to express the results of clustering, a set of features can be defined to describe the distribution of inter- and intra-clusters. The definition of these features will be elaborated in detail in Section \ref{sec:Features of combat space clustering}.

\subsubsection{Design of Scripts}
\label{sec:design_of_scripts}
Based on scripts and combat space clustering, this paper designs script sets that represent offensive, defensive, and mixed tactics. The types of targets encompass both enemy units and map points. To comprehensively assess the algorithm's performance, additional ineffective distractors such as random actions (do\_randomly) and taking no action (do\_nothing) are incorporated for comparative analysis, as detailed in Table~\ref{tab:macro-action}.

\begin{table}[htbp]
\begin{center}
\caption{Scripts designed in this paper}\label{tab:macro-action}
    \resizebox{0.42\textwidth}{!}{
    \begin{tabular}{|c|c|c|}
      \hline
      Tactic & Script & Target \\
      \hline
      \multirow{5}{*}{offensive} & ATK\_nearest & \multirow{5}{*}{unit} \\
      \cline{2-2}
      & ATK\_clu+nearest &  \\
      \cline{2-2}
      & ATK\_nearest+weakest &  \\
      \cline{2-2}
      & ATK\_clu+nearest+weakest &  \\
      \cline{2-2}
      & ATK\_threatening &  \\
      \cline{1-3}
      \multirow{1}{*}{defensive} & DEF\_clu+nearest & \multirow{2}{*}{point}  \\
      \cline{1-2}
      \multirow{3}{*}{mixed} & MIX\_gather & \\
      \cline{2-3}
      & MIX\_lure & \multirow{2}{*}{mix} \\
      \cline{2-2}
      & MIX\_sacrifice+lure & \\
      \hline
      \multirow{2}{*}{interferential} & do\_randomly & unit \\
      \cline{2-3}
      & do\_nothing & - \\
      \hline
    \end{tabular}
    }
\end{center}
\end{table}

The scripts are mainly categorized into three types: ATK (attack), DEF (defense), and MIX (mixed), which correspond to offensive, defensive, and mixed tactical intentions. For example, ATK\_threatening represents attacking threatening enemy units in the global, while DEF\_clus+nearest represents retreating from the closest enemy units in the clusters. MIX\_gather represents a mixed tactic of gathering the team in clusters, while MIX\_lure and MIX\_sacrifice+lure respectively represent mixed tactics of luring enemies and luring enemies with few units sacrifice.

\subsection{Multi-table HRL Architecture}
The paper uses Q-learning \cite{watkins_Learning_1989} to obtain optimal strategies in different situations. However, a single-table model is not able to learn the optimal strategy for different local clusters simultaneously.

\begin{figure*}[t]
    \centering
    \includegraphics[width=\linewidth]{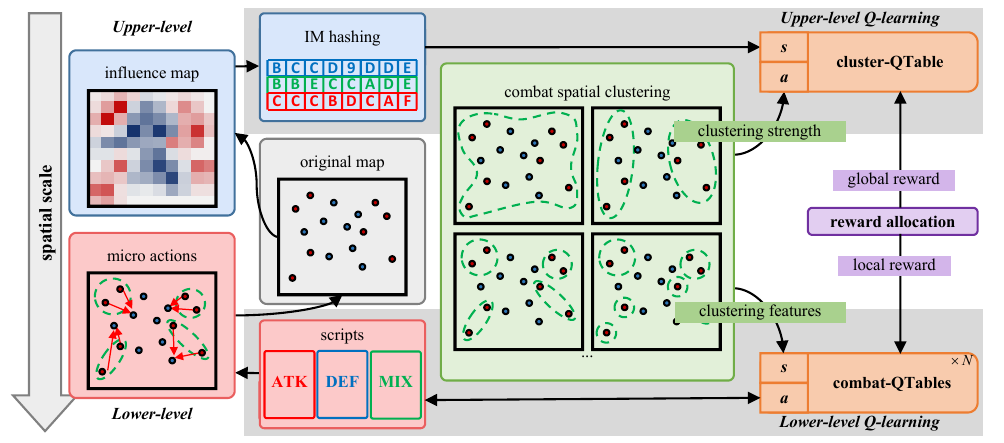}
    \caption{Illustration of hierarchical multi-table Q-learning model. The whole architecture is divided into upper and lower levels based on spatial scale. For upper level, the spatial combat information on the original map is expressed to state by influence map hashing. For lower level, the action space is formed by cluster-based scripts, which inducts micro actions. Combat spatial clustering provides clustering strength for the upper-level action space and clustering features for the lower-level multi-table index and state space. So far a hierarchical multi-table Q-learning model is formed, in which global and local rewards will be allocated.}
    \label{fig:HRL}
\end{figure*}
\label{sec:Hierarchical multi-table Q-learning model framework}
To address these issues, this paper proposes an extension of the single-table Q-learning model into a hierarchical multi-table Q-learning model. The framework of this extended model is illustrated in Figure~\ref{fig:HRL}. The upper level employs a single clustering Q-table to learn the optimal strategy on a global scale. In contrast, the lower level utilizes multiple combat Q-tables to learn optimal tactics on a local scale. These levels are interconnected through an intermediate level, which involves combat space clustering and a local reward mechanism.

\subsubsection{Upper Level}
The upper level table reflects the expected return of executing combat space clustering in various situations.  In the upper level table $\mathcal{Q}^{u}$, the state space $\mathcal{S}_{hash}$ is composed of the influence map hashing, and the action space $\mathcal{A}^{u}$ consists of a series of discrete clustering strengths $C_s$ of the combat space clustering. It should be noted that, to ensure map-scale-independent discretization, this work introduces the clustering strength parameter $C_s \in [0, 1]$ to control the ratio of the number of clusters $C_k$ to the total number of units $N$. An appropriate discrete set of $C_s$ is selected as the action space, thereby ensuring the algorithm’s scalability across heterogeneous scenarios with the cardinality of the action space kept bounded.

\subsubsection{Intermediate Level}
\label{sec:Features of combat space clustering}
Clustering features reflect the overall distribution of clusters obtained after performing clustering operations on the upper level, which will be used to establish indexes for the lower level table. After $C_s$ has been determined in the upper level, the number of clusters $C_k$ among $N$ combat units can be given as 
\begin{equation}
    C_k=\lceil N(1-C_s)+C_s \rceil,\quad C_k \in \mathbb{Z},\quad C_k \in [1, N]
\end{equation}
and the combat units can be divided into $C_k$ clusters using k-means clustering. Throughout this paper, uppercase $C_k$ denotes the number of clusters, while lowercase $c_i$ represents the $i$-th cluster as a set of units. Meanwhile, a set of inter- and intra-cluster features are defined to measure the distribution of clusters and individual units within each cluster.

\label{sec:inter-cluster features}
\label{sec:intra-cluster features}
\begin{enumerate}
\item[a)] Inter-cluster features: The inter-/intra-cluster variance ratio, also named as the Calinski-Harabasz index which is commonly used to evaluate clustering effectiveness, is used to measure cluster distribution for establishing the index of the lower table. The inter-/intra-cluster variance ratio $\mathrm{CH}(C_k)$ is defined as
    \begin{equation}
        \mathrm{CH}(C_k)=\frac{SS_B}{SS_W} \times \frac{N-C_k}{C_k-1}
    \end{equation}
where $C_k$ is the number of clusters and  $SS_B$ represents the inter-cluster variance, and $SS_W$ represents the intra-cluster variance. The inter-cluster variance $SS_B$ is defined as
\begin{equation}
    SS_B=\sum_{i=1}^{C_k}N_i\|m_i-m\|^2
\end{equation}where $N_i$ represents the number of units in cluster $i$, $m_i$ represents the centroid of cluster $i$, $m$ represents the mean of coordinates of all units, and $\|*\|$ represents the Euclidean distance between two vectors. The intra-cluster variance $SS_W$ is defined as
\begin{equation}
    SS_W=\sum_{i=1}^{C_k}\sum_{x \in c_i}\|x-m_i\|^2
\end{equation}where $x$ represents the coordinate of a unit in cluster $c_i$.

\item[b)] Intra-cluster features: This feature includes the radial dispersion $d_r$ and circumferential dispersion $d_c$. $d_r$ is calculated by the coefficient of variation of the distance between the coordinate of each unit in a cluster and the centroid of the cluster, which represents the dispersion of units radially within the cluster. The radial dispersion $d_{r,i}$ of cluster $i$ is defined as
\begin{equation}
    \label{eq:radial_dispersion}
    d_{r,i}=\frac{\sigma_{r,i}}{\mu_{r,i}}=\frac{\sqrt{\frac{1}{N_i} \sum_{j=1}^{N_i} (d_{i,j} - \mu_{r,i})^2}}{\frac{1}{N_i} \sum_{j=1}^{N_i} d_{i,j}}
\end{equation}where $\sigma_{r,i}$ represents the standard deviation of the distance between units and the centroid of cluster $i$, $\mu_{r,i}$ represents the mean distance between units and the centroid of cluster $i$, $N_i$ represents the number of units in cluster $i$, and $d_{i,j}$ is the distance between $m_i$ and unit $j$ in cluster $i$. The circumferential dispersion $d_c$ is calculated by the coefficient of variation of the distance between each unit within a cluster and its nearest neighbor unit, which represents the dispersion of units in the circumferential direction within the cluster. The circumferential dispersion $d_{c,i}$ of cluster $i$ is defined as
\begin{equation}
    \label{eq:circumferential_dispersion}
    d_{c,i}=\frac{\sigma_{c,i}}{\mu_{c,i}}=\frac{\sqrt{\frac{1}{N_i} \sum_{j=1}^{N_i} (d'_{i,j} - \mu_{c,i})^2}}{\frac{1}{N_i} \sum_{j=1}^{N_i} d'_{i,j}}
\end{equation}where $\sigma_{c,i}$ represents the standard deviation of the distance between units and their nearest neighbor unit in cluster $i$, $\mu_{c,i}$ represents the mean distance between units and their nearest neighbor unit in cluster $i$, $N_i$ represents the number of units in cluster $i$, and $d'_{i,j}$ represents the distance between unit $j$ in cluster $i$ and its nearest neighbor unit.
\end{enumerate}

In conclusion, $\mathrm{CH}(C_k)$ and $C_k$ index which Q-table to consult, while $d_{r,i}$ and $d_{c,i}$ provide the state for the lower level.

\subsubsection{Lower Level}
Multiple lower level tables are created based on the different inter-cluster characteristics formed by each combat space clustering. Each lower level table reflects the expected return of units in each cluster when executing scripts in the combat scenario. In lower level tables $\mathcal{Q}^{l}$, the state space $\mathcal{S}^{l}$ is composed of the intra-cluster features, and the action space consists of scripts from the shared script set $\Omega$. To obtain discrete states in the lower-level Q-table, the continuous intra-cluster features is discretized into a two-symbol hexadecimal key as follows:
\begin{enumerate}
	\item[a)] Extract features: $f_{r,i}=1-d_{r,i}$, $f_{c,i}=1-d_{c,i}$, where $d_{r,i}$ and $d_{c,i}$ are the radial and circumferential dispersions computed from Eq.~\eqref{eq:radial_dispersion} and Eq.~\eqref{eq:circumferential_dispersion}, respectively.
	\item[b)] 4-bit quantization: $F_{r,i}=\min\bigl(15,\lfloor 15.9\,f_{r,i}\rfloor\bigr)$, \ $F_{c,i}=\min\bigl(15,\lfloor 15.9\,f_{c,i}\rfloor\bigr)$, where the factor $15.9$ guarantees that the maximum value $1.0$ is mapped to the largest 4-bit integer $15$ without overflow.
	\item[c)] Hex encoding: $s_i=\texttt{\small \{:01X\}\{:01X\}.format}(F_{r,i},F_{c,i})$, or \texttt{\small X} if the cluster set is empty, where each integer $F_{r,i}$ and $F_{c,i}$ is converted into a single hexadecimal digit \texttt{0-F}, and the special symbol \texttt{\small X} is used when the cluster set is empty.
\end{enumerate}
Through the above processing, the clustering features are linearly rescaled to the discrete space and concatenated into a hexadecimal string, serving as the discrete state in the Q-table.

\subsection{Local Reward Mechanism}

The upper and lower level Q-tables represent the strategies at the global and local levels, respectively. Therefore, they require different reward mechanisms. For the upper level, the global reward at $t$ time can be directly derived from the difference between the damage inflicted by self-units and the damage sustained, encapsulating the overall gain of an action, which can be represented as
\begin{equation}
    \begin{aligned}          r_{g}\left(t\right)&=H_s\left(t\right)-H_e\left(t\right)
    \end{aligned}
\end{equation}
where $H_s\left(t\right)$ represents the hitpoint of self units at $t$ time and $H_e\left(t\right)$ represents the enemy's. 

However, for the lower level, each cluster may execute different scripts or demonstrate different tactics, so the local reward $r_{l,i}\left(t\right)$ for cluster $i$ is designed as
\begin{equation}
    \begin{aligned}          r_{l,i}\left(t\right)&=r_{enemyloss}\left(t\right)+r_{selfloss}\left(t\right)\\
    &+r_{enemykilled}\left(t\right)+r_{selfkilled}\left(t\right)\\
    &+r_{comeback}\left(t\right)+r_{setback}\left(t\right)     
    \end{aligned}
\end{equation}
where $r_{enemyloss}\left(t\right)$ represents the score corresponding to the rate of enemy health loss. For every 1\% of enemy health lost, this item gains 1 point. Reward $r_{selfloss}\left(t\right)$ represents the score corresponding to the rate of self-health loss. For every 1\% of self-health lost, this item deducts 1 point. Reward $r_{enemykilled}\left(t\right)$ represents the score corresponding to the elimination of enemy units. For every 1 enemy unit eliminated, this item gains 5 points. Reward $r_{selfkilled}\left(t\right)$ represents the score corresponding to the elimination of self-units. For every 1 self-unit eliminated, this item deducts 5 points. Reward $r_{comeback}\left(t\right)$ is set to 10 when the self-side transitions from a disadvantaged situation to an advantage. Reward $r_{setback}\left(t\right)$ is set to -10 when the self-side transitions from an advantageous situation to a disadvantage. Note that the values are set based on preliminary studies.

\subsection{Training Algorithm}
The complete training procedure for HRL-IM/CBS is presented in Algorithm~\ref{alg:training}. The algorithm alternates between upper-level and lower-level Q-value updates, with the hierarchical structure enabling efficient credit assignment across different spatial scales.

The training algorithm exhibits two key properties that contribute to its efficiency. The hierarchical structure enables separate learning of global clustering decisions and local script selections, effectively reducing the action space complexity at each decision level. Additionally, the local reward mechanism $r_{l,i}$ enables direct attribution of outcomes to specific cluster actions, addressing the multi-agent credit assignment problem. Together with the tabular Q-learning approach that avoids the computational overhead of neural network forward/backward propagation, these properties enable lightweight and sample-efficient training compared to end-to-end DRL methods.

\renewcommand{\algorithmcfname}{Algorithm}
\begin{algorithm}
    
	\caption{HRL-IM/CBS Training}\label{alg:training}
	\renewcommand{\thealgocf}{}
	\SetAlgoLined
	\KwIn{Number of episodes $E$, max steps per episode $K$, learning rate $\alpha$, discount factor $\gamma$, exploration schedule $\varepsilon(t)$}
	\KwIn{Script set $\Omega$, clustering strength set $\mathcal{C}_s$}
	\KwOut{Trained Q-tables $\mathcal{Q}^{u}$ and $\mathcal{Q}^{l}$}

	Initialize $\mathcal{Q}^{u}(s, C_s) \leftarrow 0$ for all $s, C_s$\;
	Initialize $\mathcal{Q}^{l}_i(s_i, \omega) \leftarrow 0$ for all clusters $i$, states $s_i$, scripts $\omega$\;

	\For{episode $e = 1$ to $E$}{
	Initialize environment, obtain initial observation $O_0$\;
	$s_0 \leftarrow \text{encode\_IM\_matrix}(O_0)$\;

	\For{step $k = 0$ to $K$}{
		\tcp{Upper-level: Select clustering strength}
		$C_s \leftarrow \varepsilon\text{-greedy}(\mathcal{Q}^{u}(s_k, \cdot), \varepsilon(e))$\;
		$C_k \leftarrow \lceil N(1-C_s)+C_s \rceil$\;
		$\{c_1, \ldots, c_{C_k}\} \leftarrow \text{k-means}(\text{unit positions}, C_k)$\;

		\tcp{Intermediate: Compute clustering features}
		Compute $\mathrm{CH}(C_k)$, $d_{r,i}$, $d_{c,i}$ for each cluster $i$\;
		$s_i \leftarrow \text{discretize}(d_{r,i}, d_{c,i})$ for each cluster $i$\;

		\tcp{Lower-level: Select scripts for each cluster}
		\For{cluster $i = 1$ to $C_k$}{
			$\omega_i \leftarrow \varepsilon\text{-greedy}(\mathcal{Q}^{l}_i(s_i, \cdot), \varepsilon(e))$\;
				Execute script $\omega_i$ for units in cluster $i$\;
				}

				\tcp{Observe rewards and next state}
				Observe global reward $r_g$ and local rewards $\{r_{l,1}, \ldots, r_{l,C_k}\}$\;
				Obtain next observation $O_{k+1}$, encode $s_{k+1}$\;

				\tcp{Update Q-values according to Eq.~\eqref{eq:upper_q} and Eq.~\eqref{eq:lower_q}}
                Update $\mathcal{Q}^{u}(s_k, C_s)$ according to Eq.~\eqref{eq:upper_q}\;
                
                \For{cluster $i = 1$ to $C_k$}{
                    Update $\mathcal{Q}^{l}_i(s_i, \omega_i)$ according to Eq.~\eqref{eq:lower_q}\;
                }

			$s_k \leftarrow s_{k+1}$\;
		}
	}
	\Return $\mathcal{Q}^{u}$, $\mathcal{Q}^{l}$\;
\end{algorithm}

\section{Experimental Setup and Design}
\label{Experimental Setup and Design}
This paper uses StarCraft II to demonstrate the effectiveness of the proposed HRL-IM/CBS method by using PySC2\footnote{https://github.com/google-deepmind/pysc2}. In this chapter, the experimental scenarios and parameter settings will be discussed.

\subsection{Experimental Scenarios}
\label{sec:Experimental scenarios in StarCraft II micro-management}

In StarCraft II, units of different races usually have different attributes, such as max hitpoints, cooldown, movement speed, damage per hit, fire range, and sight range. Some previous studies have assigned different types of units to opposing sides in experimental setups, which may grant certain unit types inherent advantages over others. This can make some strategies highly adaptive in specific scenarios. For instance, \textquotedblleft kiting\textquotedblright{ }tactics that leverage movement speed can easily triumph even when outnumbered. Other research has focused on mixed team with multi unit types, which is an effective way to ensure scenario complexity and experimental fairness. However, the overall strength of mixed team cannot be simply assessed by the number of units and their hitpoints. Determining the composition ratio and overall strength of mixed team is challenging, which affects the design of the influence map.

In this paper, mixed teams are not considered for now, and instead focus on creating fair scenarios with equal numbers and types of units for both sides. This setup aims to ensure comparable levels of both strength, which also means those strategies that rely on movement speed and attack range to easily win may not work, and poses a greater challenge for algorithms. The Marine unit type is uniformly used, and the attributes of the unit are shown in Table~\ref{tab:unit_params}.

\begin{table}[ht]
\vspace{-5pt}
    \caption{Attributes of Marine unit type}\label{tab:unit_params}
    \centering
    \begin{tabular}{cccccc}
      \hline
      Unit type & Max hitpoint & Fire range & Damage & Cooldown\\
      \hline
      Marine & 45 & 5 & 6 & 15\\
      \hline
    \end{tabular}
\vspace{-5pt}
\end{table}

\begin{figure}[ht]
	\centering
	\begin{minipage}{0.3\linewidth}
		\centering
		\includegraphics[width=\linewidth]{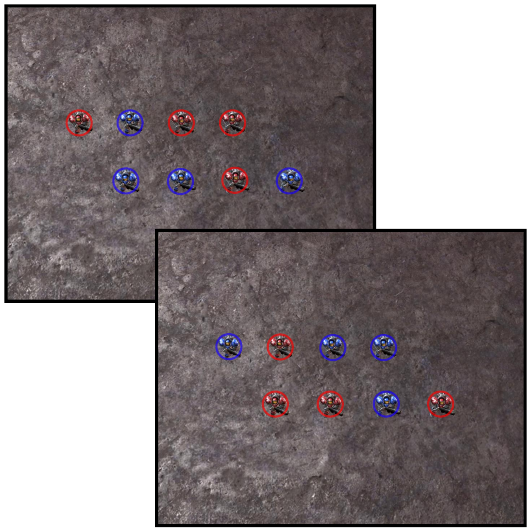}
		a) sce-1\&1m
	\end{minipage}
	\begin{minipage}{0.3\linewidth}
		\centering
		\includegraphics[width=\linewidth]{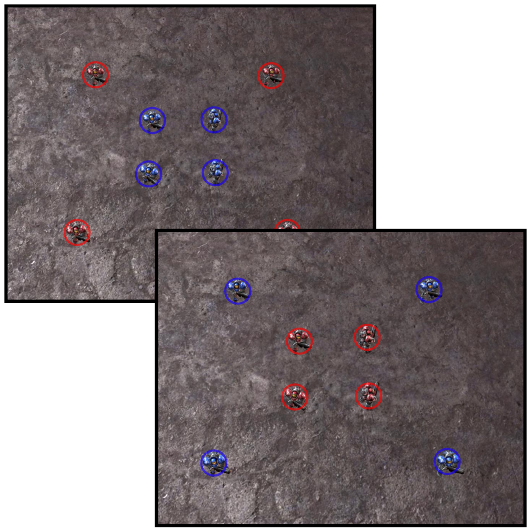}
		b) sce-2\&2m
	\end{minipage}
	\begin{minipage}{0.3\linewidth}
		\centering
		\includegraphics[width=\linewidth]{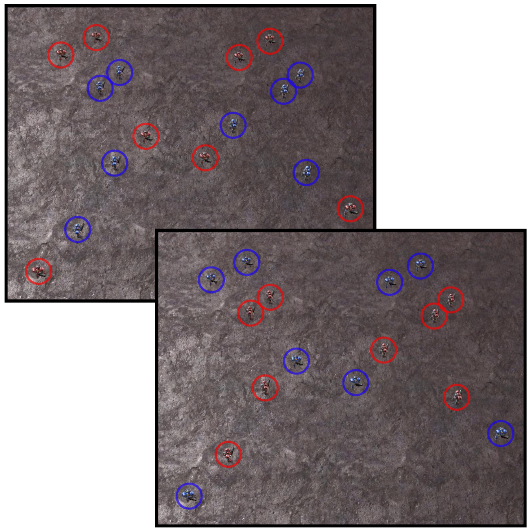}
		c) sce-3\&3m
	\end{minipage}
	\caption{Unit distributions of experimental scenarios. Different colors represent opposing forces, and distributions are swapped in mirror scenarios to ensure positional fairness.}
	\label{fig:scenarios}
\end{figure}

Additionally, formations within and between teams are also critical factors that have often been overlooked in previous experiments. In many cases, the two opposing sides were simply placed on opposite ends of the map, positioned at such a distance that it tended to obscure the direct interaction and combat tension, thereby undermined the immediacy of their confrontational dynamics. In this paper, the impact of interspersed unit distribution in equal confrontation has been considered. Taking into account the characteristics of formations and strategies, a series of experimental scenarios have been designed, as illustrated in Figure~\ref{fig:scenarios} and described in Table~\ref{tab:scenarios}.

\begin{table}[ht]
    \caption{Descriptions of experiment scenarios designed in this paper}\label{tab:scenarios}
    \centering
    \begin{tabular}{cp{0.78\linewidth}}
        \hline
      Scenario & Description\\
        \hline
        sce-1 & Both sides have 4 Marines with the same troop configurations. Additionally, units aggregation and isolation have been considered.\\
        sce-1m & The mirror map of sce-1. Both sides have 4 Marines with the same troop configurations. The map distribution is similar to sce-1, with the only difference being the positions of both sides are swapped, so as to verify the effectiveness of the algorithm rather than relying on advantages brought by the map distribution.\\\hline
        sce-2 & Both sides have 4 Marines with the same troop configurations but different formation. Self units are disperse but surrounding enemy units on the map.\\
        sce-2m & The mirror map of sce-2.\\\hline
        sce-3 & Both sides have 8 Marines with the same troop configurations. The unit layout is asymmetrical, scattered and disorder, which has greater complexity than sce-1 and usually requires longer training time to ensure convergence.\\
        sce-3m & The mirror map of sce-3.\\\hline
    \end{tabular}
\end{table}

In the experimental setup, the self-agent employs the HRL-IM/CBS strategy proposed in this paper, while the enemy utilizes the built-in strategy. The two sides engage in combat on designated maps with initial formations until one side is entirely eliminated. The outcomes of each match, including win/loss results and scores, are recorded. Furthermore, the proposed method was compared with DRL baselines (QMIX \cite{rashid_Monotonic_2020}, QTRAN \cite{son_QTRAN_2019}, IQL \cite{tampuu_Multiagent_2017}, COMA \cite{foerster_counterfactual_2018} and VDN \cite{sunehag_ValueDecomposition_2017}) on the PyMARL\footnote{https://github.com/oxwhirl/pymarl}, and MAPPO\cite{yu_Surprising_2022}, which is implemented by marlbenchmark\footnote{https://github.com/marlbenchmark/on-policy} and has achieved state-of-the-art results in cooperative multi-agent games. To ensure a fair comparison, the API for the implementation of all the selected comparison algorithms is SMAC\footnote{https://github.com/oxwhirl/smac}, and every algorithm followed the same experimental pipeline and evaluation protocol. For MAPPO, all hyper-parameters (rollout threads, episode length, PPO epoch, mini-batches, clip term, etc.) were kept strictly as specified in the authors’ official repository.

\begin{figure*}[th]
    \centering
    
    \begin{minipage}{0.32\textwidth}
        \centering
        \includegraphics[width=\columnwidth]{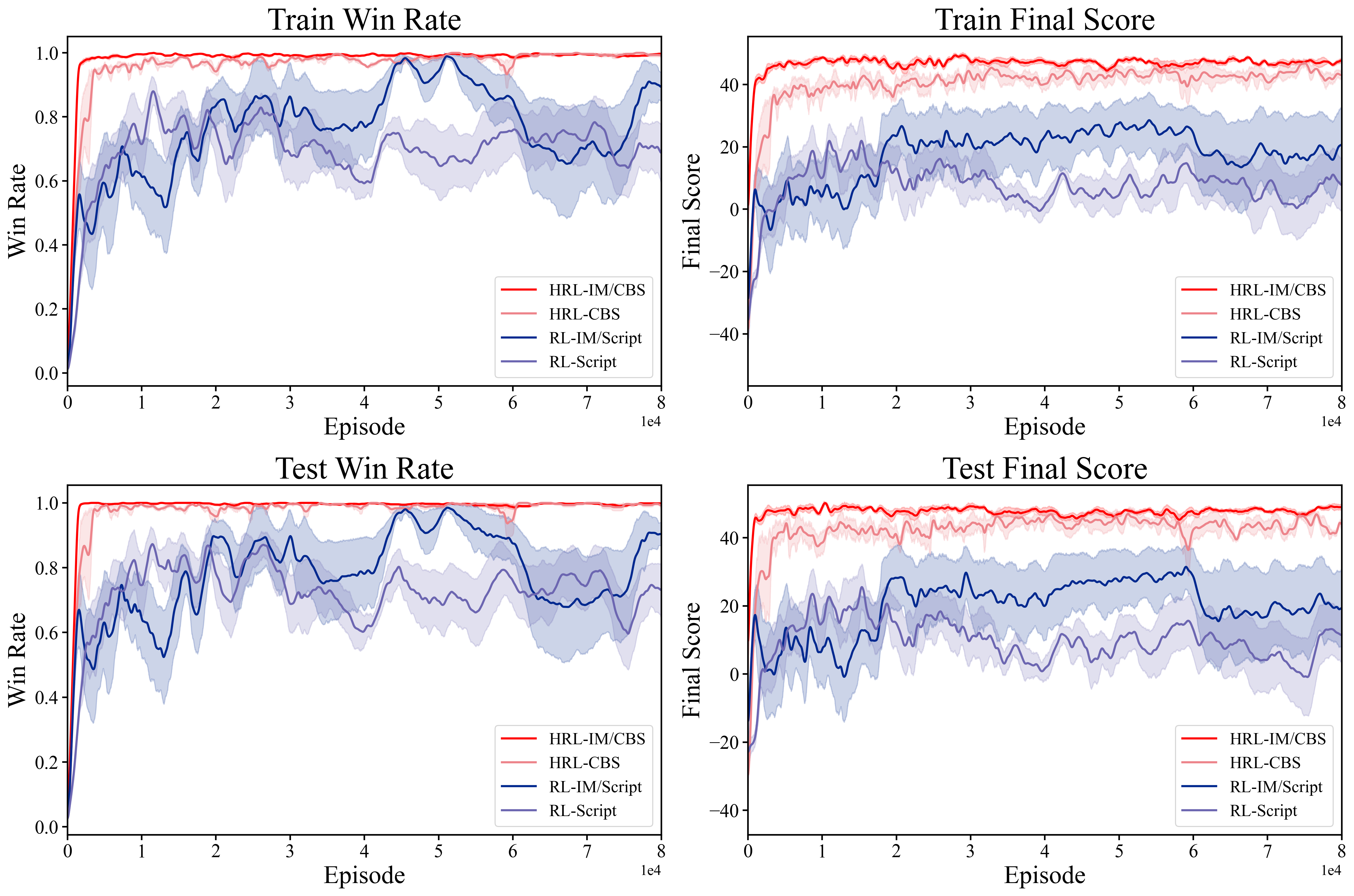}
        a) sce-1
    \end{minipage}
    \begin{minipage}{0.32\textwidth}
        \centering
        \includegraphics[width=\columnwidth]{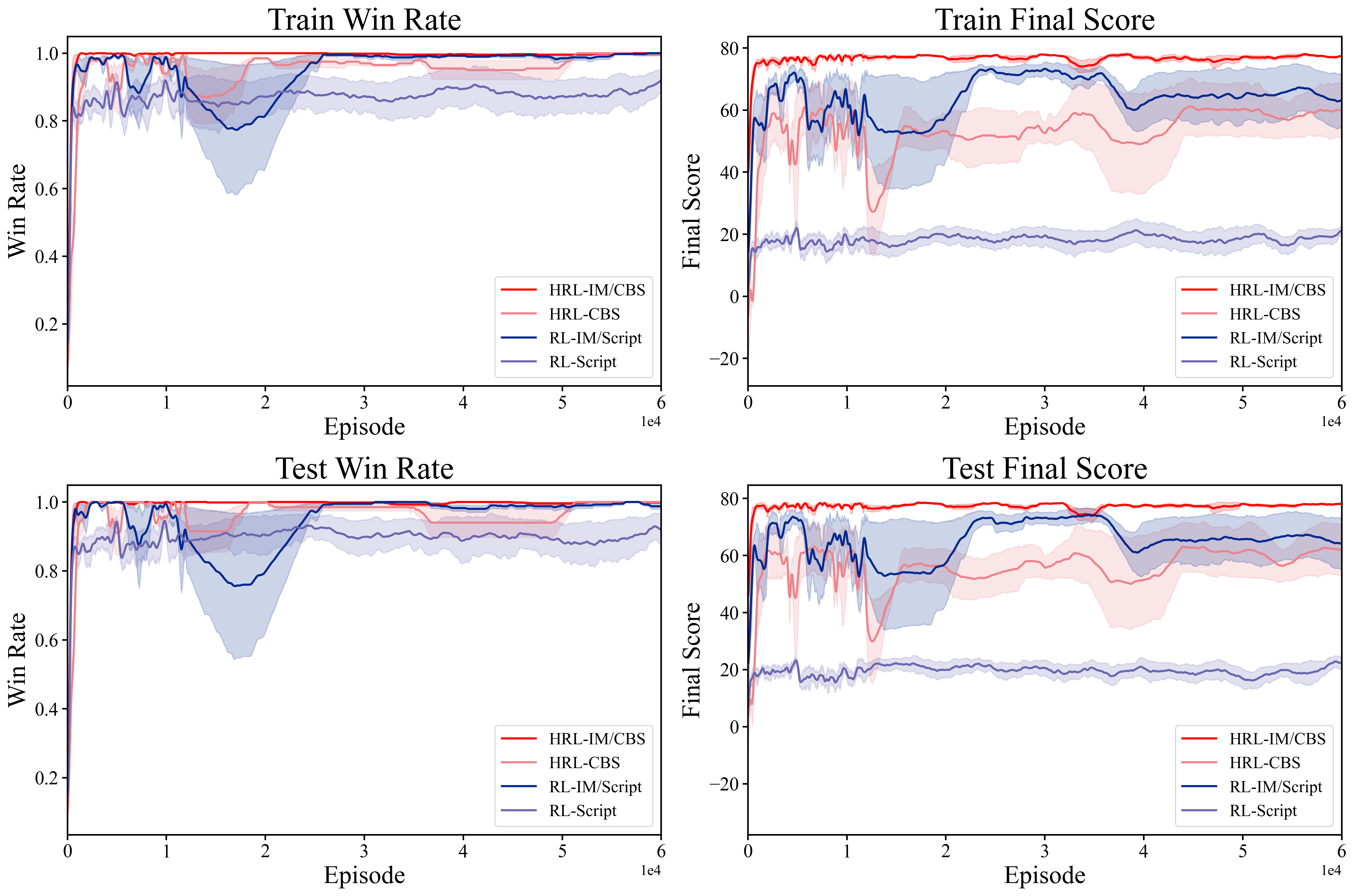}
        b) sce-2
    \end{minipage}
    \begin{minipage}{0.32\textwidth}
        \centering
        \includegraphics[width=\columnwidth]{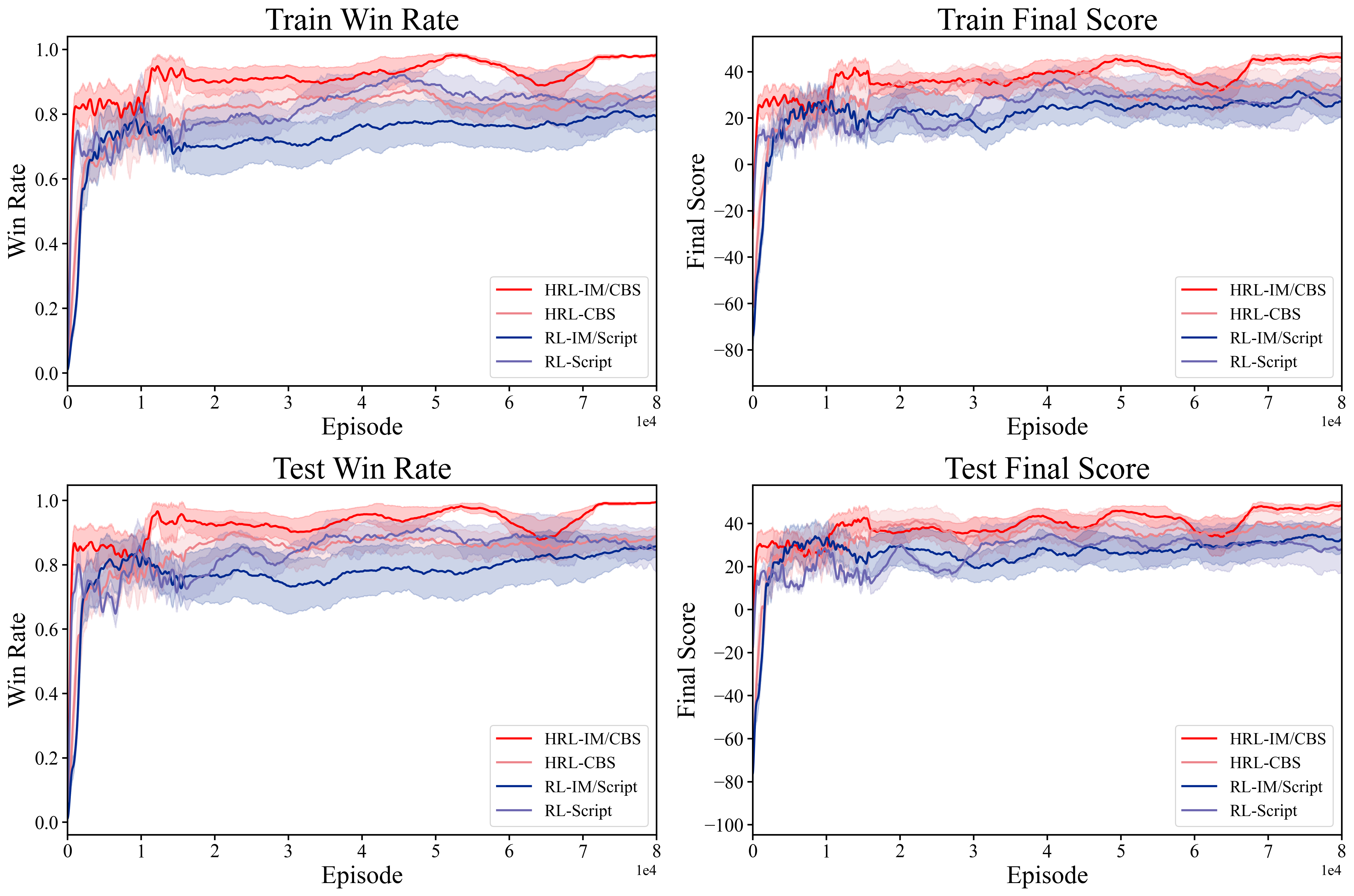}
        c) sce-3
    \end{minipage}

    \vspace{0.3cm}

    \begin{minipage}{0.32\textwidth}
        \centering
        \includegraphics[width=\columnwidth]{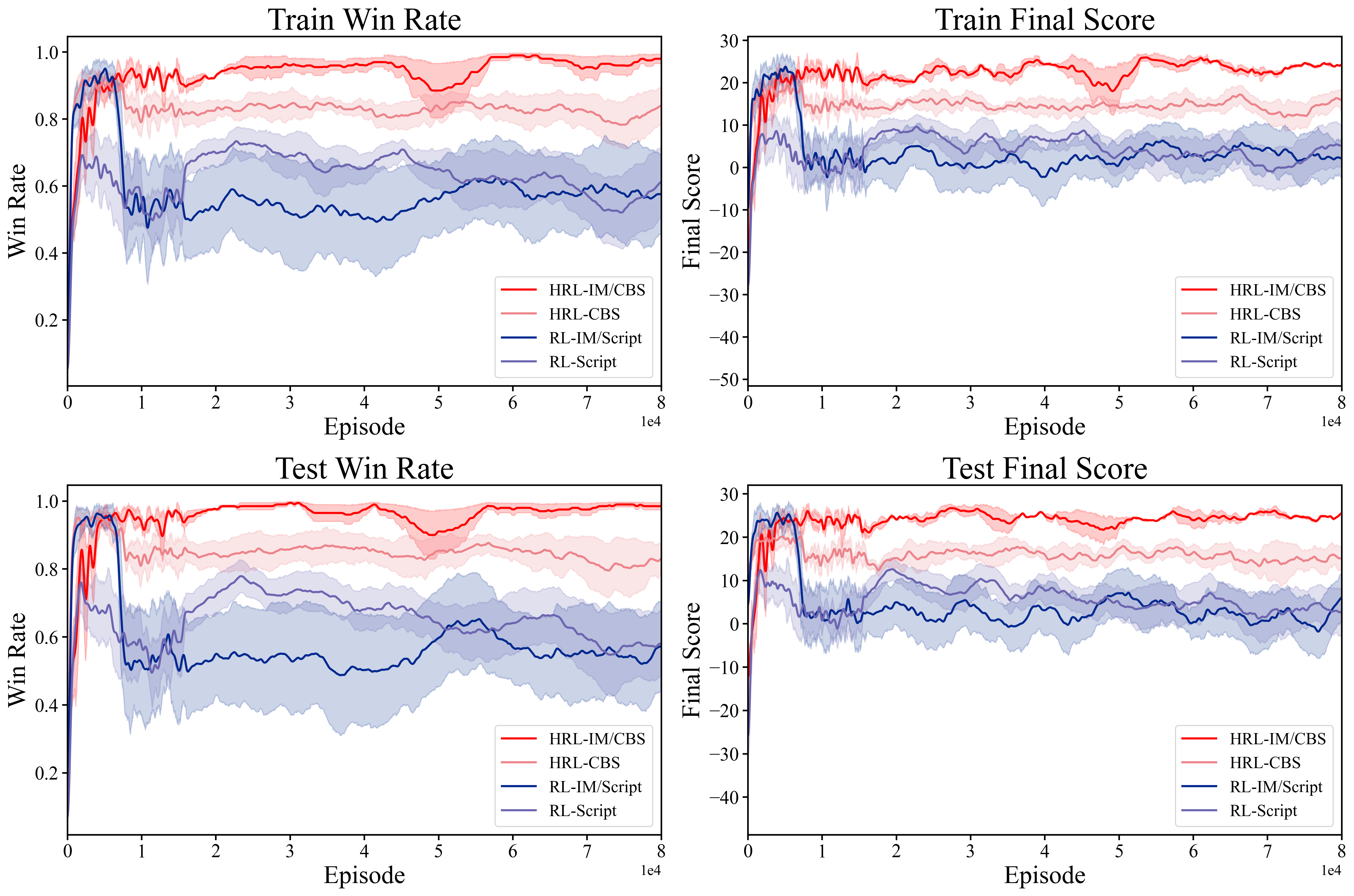}
        d) sce-1m
    \end{minipage}
    \begin{minipage}{0.32\textwidth}
        \centering
        \includegraphics[width=\columnwidth]{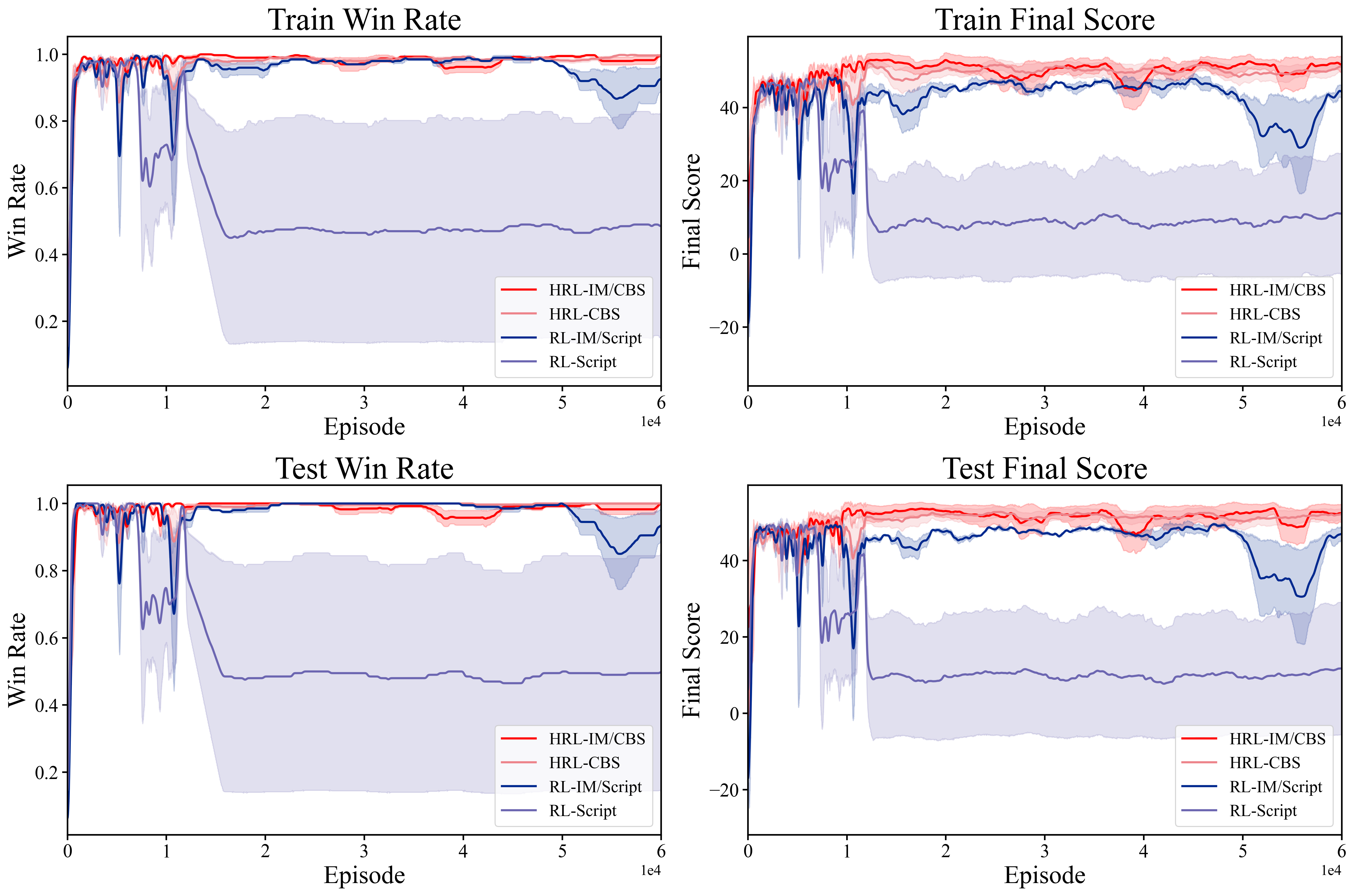}
        e) sce-2m
    \end{minipage}
    \begin{minipage}{0.32\textwidth}
        \centering
        \includegraphics[width=\columnwidth]{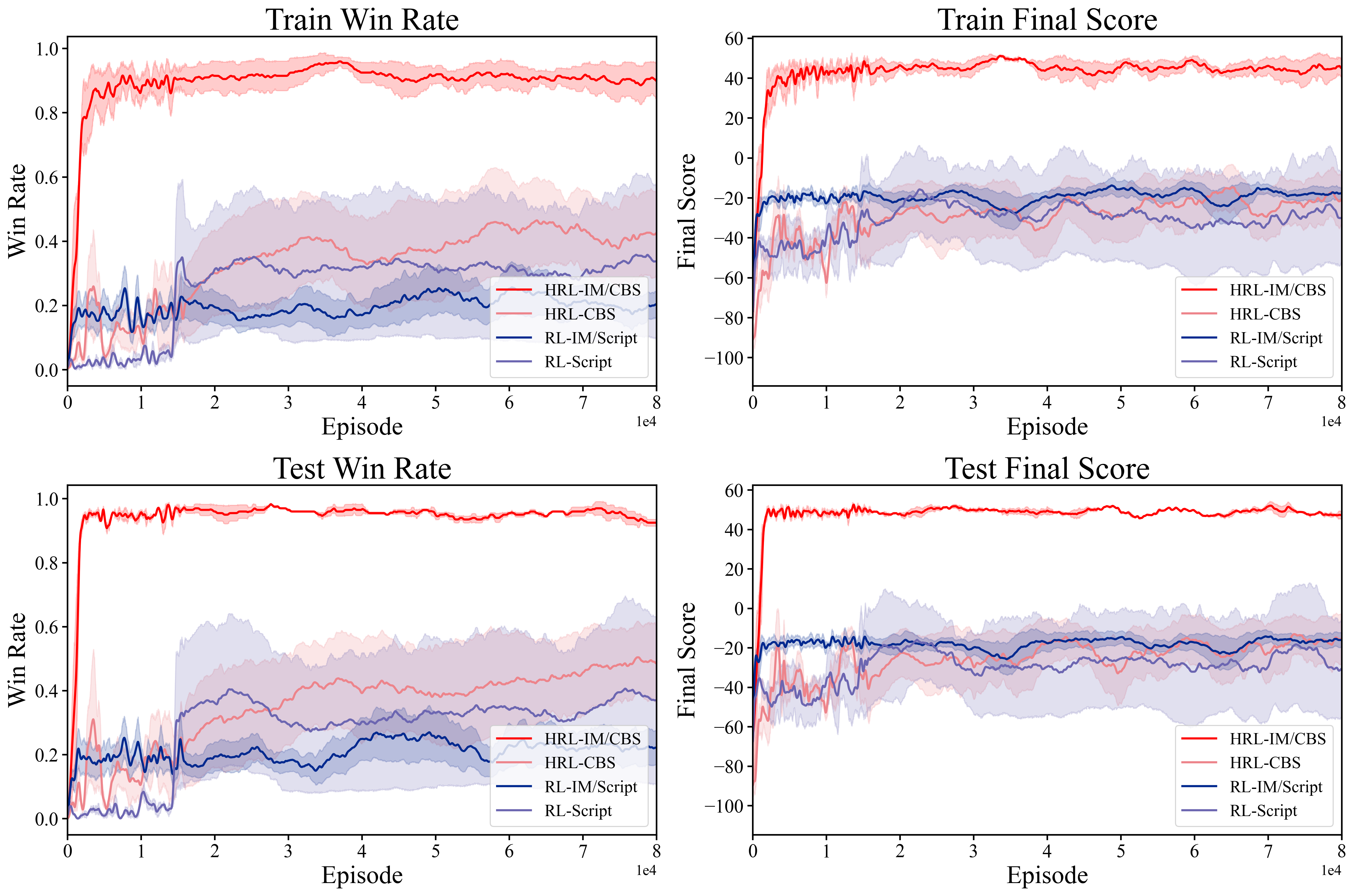}
        f) sce-3m
    \end{minipage}
    
\caption{Component ablation study of HRL-IM/CBS across all scenarios.}
\label{fig:ablation}
\end{figure*}

\subsection{Parameter Settings}
\subsubsection{Frame Skip Parameter}
For real-time games, it is impractical to make decisions frame by frame. A commonly used method is to use frame skip, which means observing and making decisions every fixed number of frames. Setting the frame skip parameter too high will reduce the effectiveness of decision-making and training, while too low will result in a sharp increase in training costs. Through experimental analysis, the frame skip parameter is set to 10, essentially causing agents to observe and make decisions every 10 frames.

\subsubsection{\texorpdfstring{$\alpha$, $\gamma$ parameter}{Alpha, Gamma Parameter}}

In the HRL-IM/CBS framework, the hierarchical Q-learning algorithm is employed with update rules formalized in Eq.~\eqref{eq:upper_q} and Eq.~\eqref{eq:lower_q}. The learning rate $\alpha$ determines how much newly acquired information overrides existing information, while the discount factor $\gamma$ determines the significance of future rewards. Through parameter comparison experiments, $\alpha$ is set to 0.1 and $\gamma$ to 0.9 to ensure effective training.

\subsubsection{\texorpdfstring{$\varepsilon$}\text{-greedy Parameter}}
In order to balance the exploration and exploitation of the Q-learning algorithm in the early and later stages of training, this paper adopts the {$\varepsilon$}\text{-greedy} mechanism, a commonly used greedy strategy. The parameter {$\varepsilon$} decreases from 0.5 to 0.05 with the number of training iterations. The formula is expressed as
\begin{equation}
	 \varepsilon=\max(0.5/\sqrt{t},0.05)
  \label{equation7}
\end{equation}
where $t$ represents the number of training iterations and $t =1,2,\dots,T.$

\section{Experimental Results and Analysis}
\label{Experimental Results and Analysis}
This section discusses and analyzes the experimental results of ablation study, comparison experiments, as well as the effectiveness and mechanism analysis of each component.

\subsection{Ablation Study}

\begin{figure*}[!th]
    \centering
    
    \begin{minipage}{0.32\textwidth}
        \centering
        \includegraphics[width=\columnwidth]{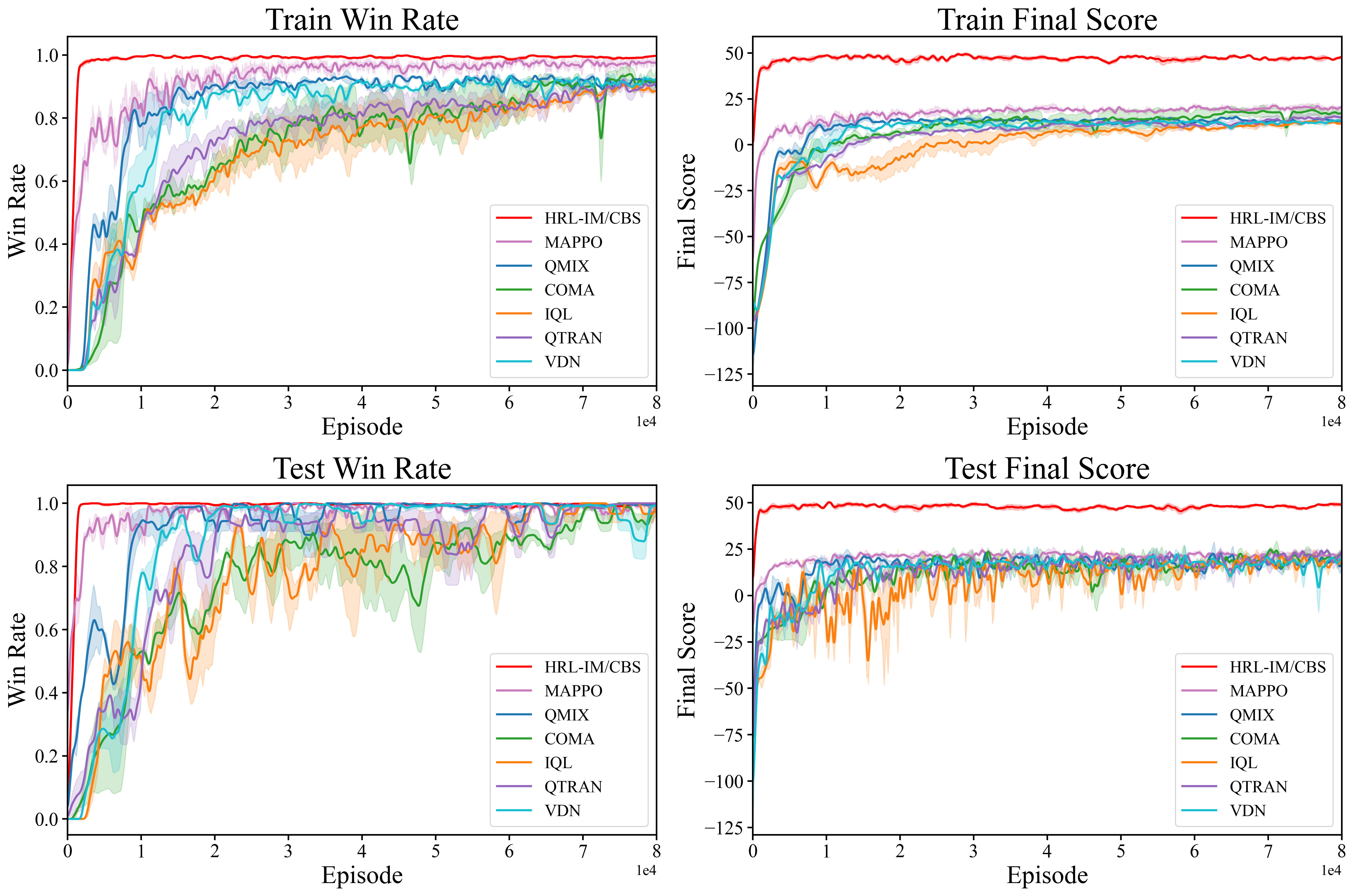}
        a) sce-1
    \end{minipage}
    \begin{minipage}{0.32\textwidth}
        \centering
        \includegraphics[width=\columnwidth]{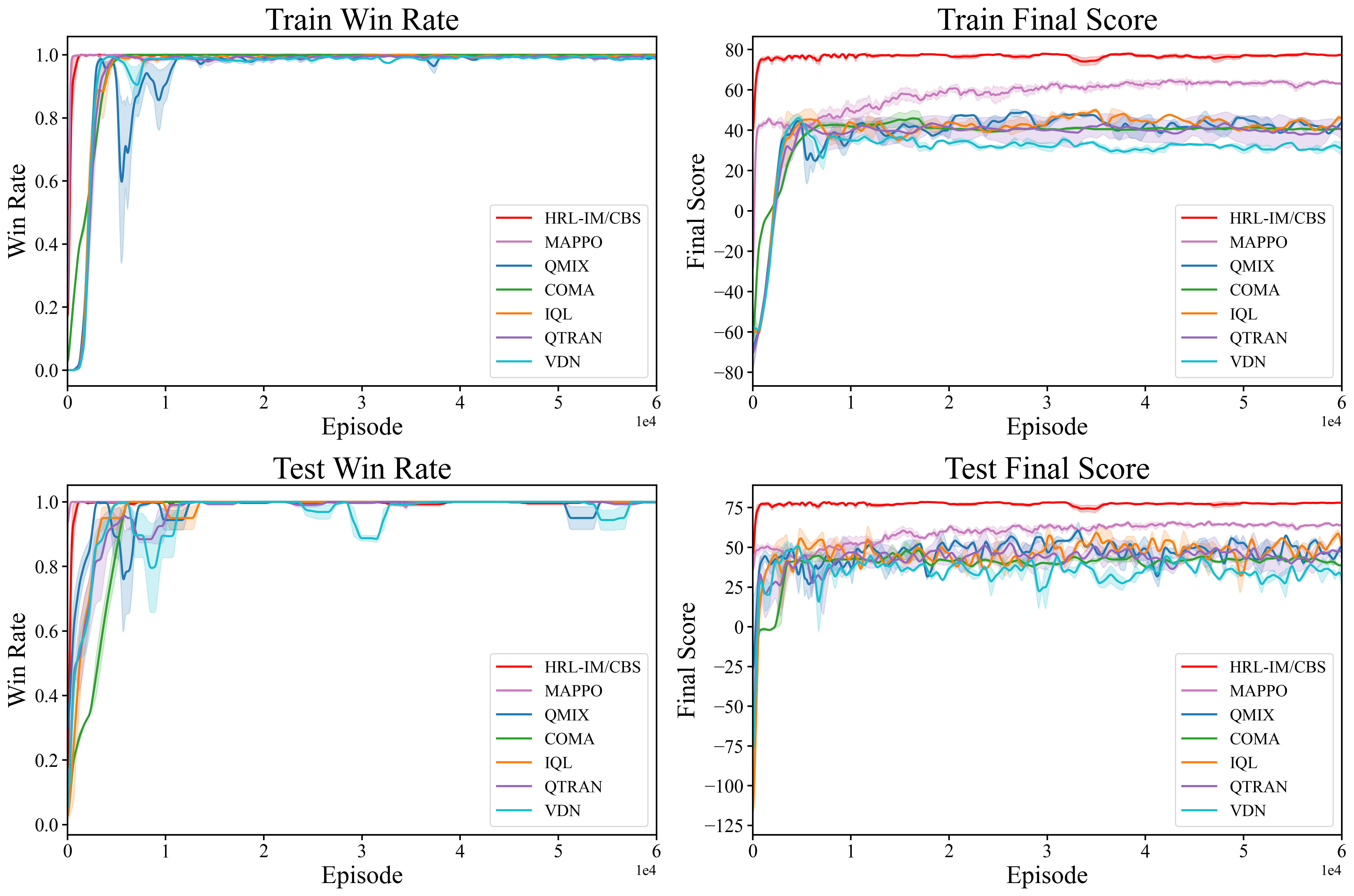}
        b) sce-2
    \end{minipage}
    \begin{minipage}{0.32\textwidth}
        \centering
        \includegraphics[width=\columnwidth]{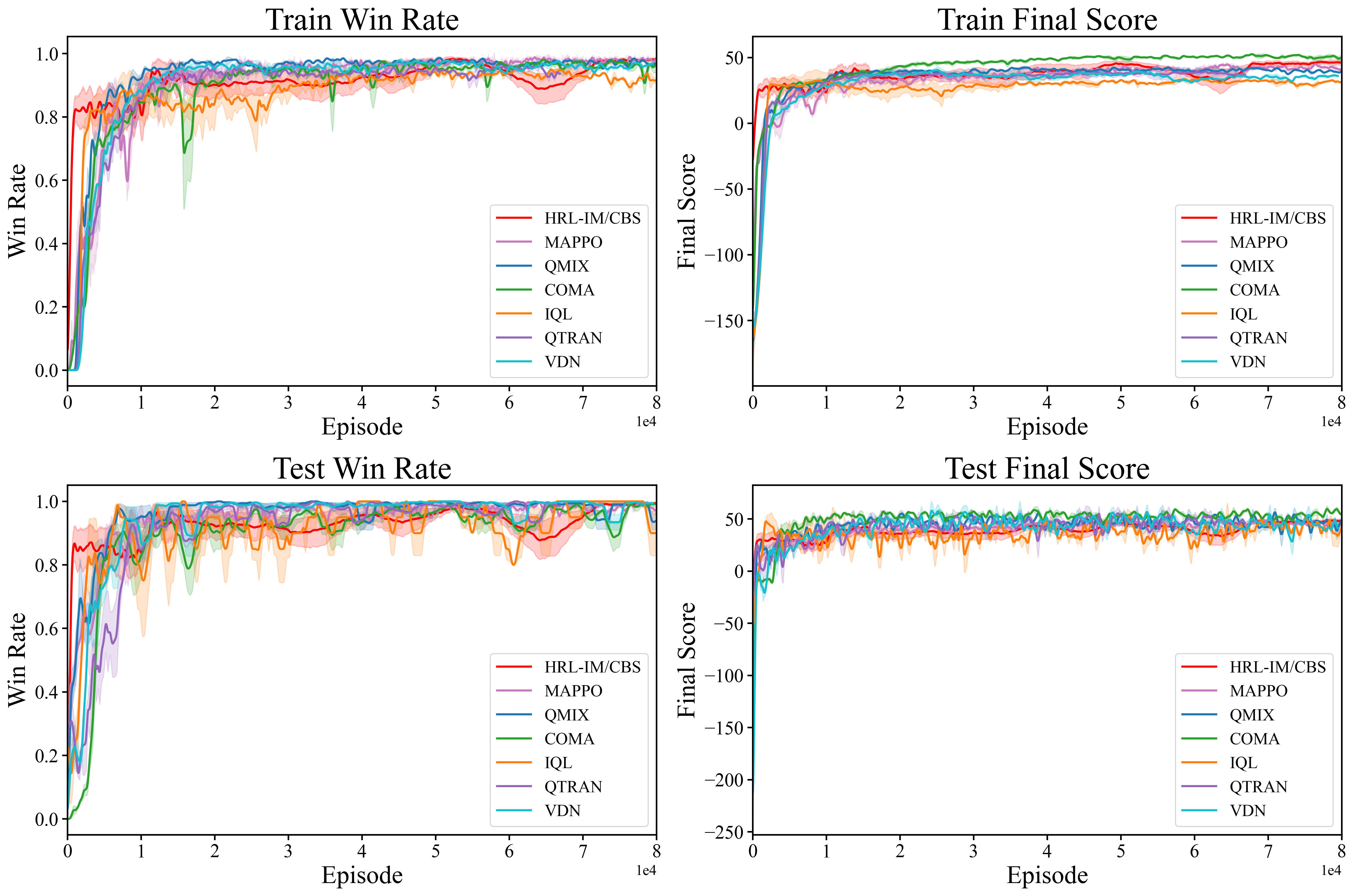}
        c) sce-3
    \end{minipage}

    \begin{minipage}{0.32\textwidth}
        \centering
        \includegraphics[width=\columnwidth]{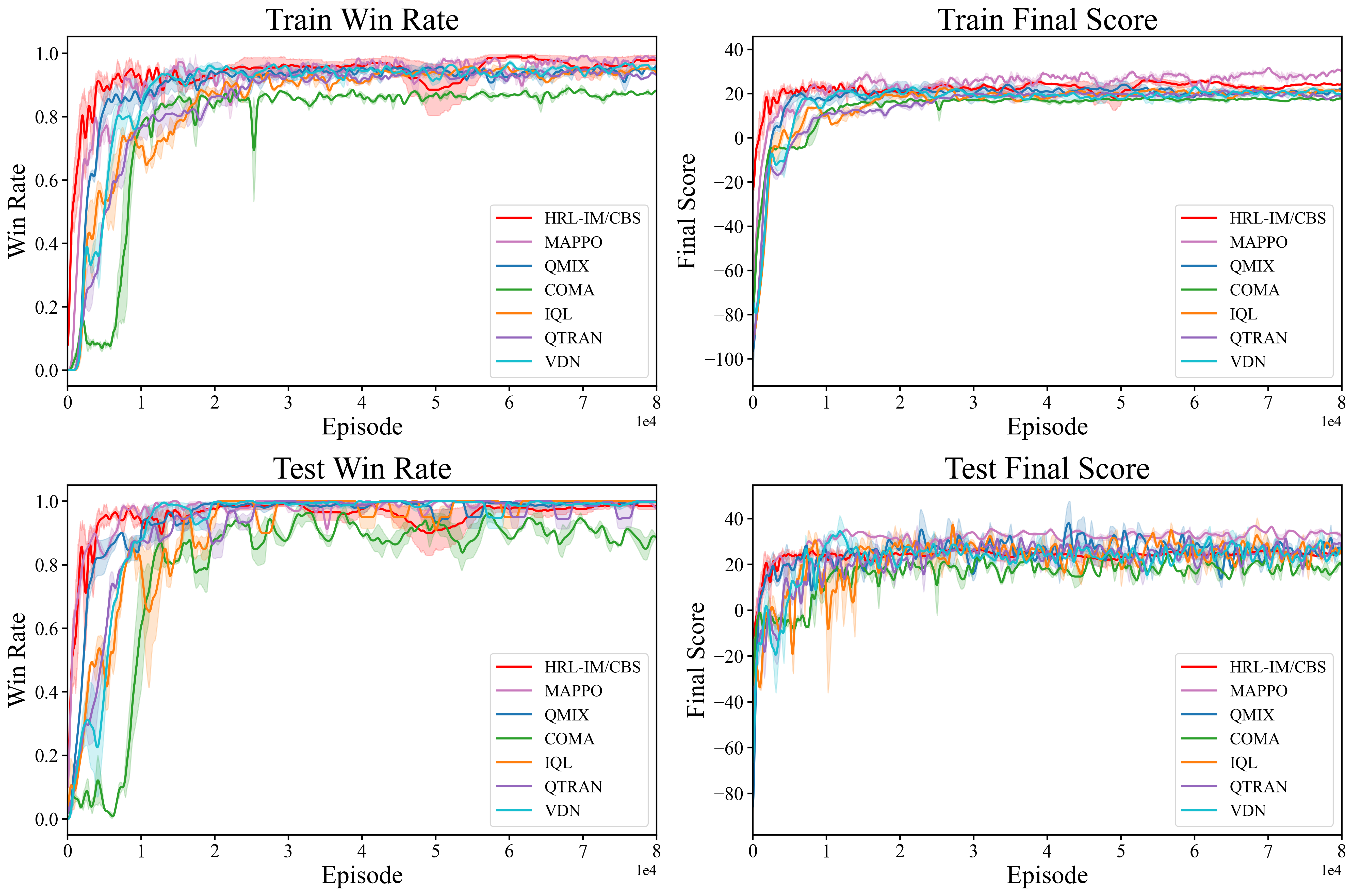}
        d) sce-1m
    \end{minipage}
    \begin{minipage}{0.32\textwidth}
        \centering
        \includegraphics[width=\columnwidth]{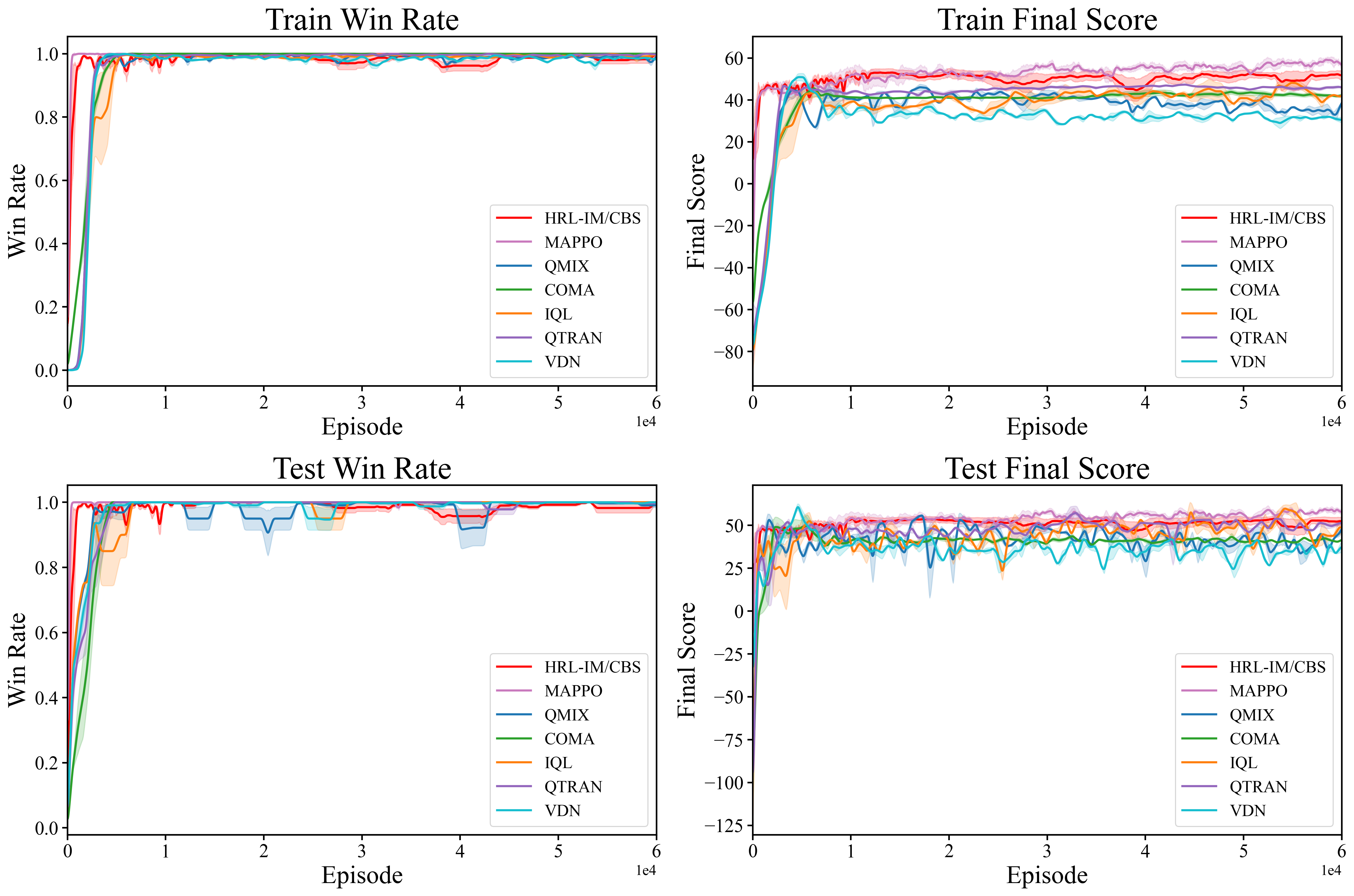}
        e) sce-2m
    \end{minipage}
    \begin{minipage}{0.32\textwidth}
        \centering
        \includegraphics[width=\columnwidth]{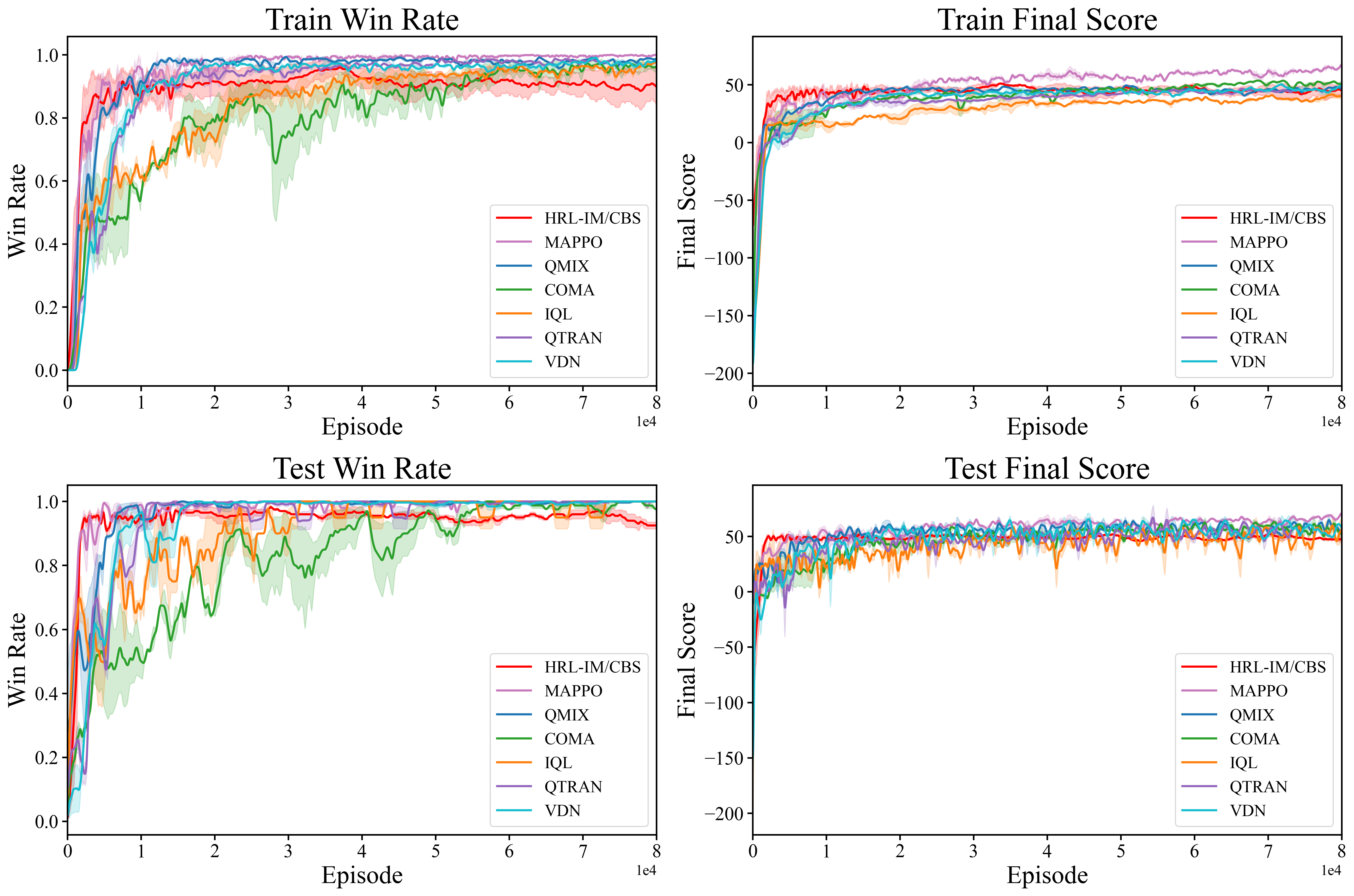}
        f) sce-3m
    \end{minipage}
    
\caption{Comparison of win rate and final score for different algorithms across all scenarios.}
\label{fig:comparison}
\end{figure*}

To rigorously assess the individual contributions of each heuristic component, a comprehensive ablation study has been conducted on HRL-IM/CBS. The ablation study evaluates three stripped-down counterparts. HRL-CBS keeps the two-level hierarchy and cluster-based scripts yet drops the influence map hashing. RL-IM/Script preserves the hash while collapsing the hierarchy into a single RL layer that commands the whole team with global scripts. RL-Script removes both enhancements and leaves a flat Q-learner that chooses scripts only from raw unit coordinates and hit-points. Quantitative results including the win rate and the final score are summarized in Table~A.~1 in Appendix A and illustrated in Figure~\ref{fig:ablation}.

The exhaustive ablation study reveals that the influence map hashing markedly boosts win rate and final score across all maps, yet HRL-CBS performs poorly in the complex sce-3m scenario. RL-IM/Script, which discards combat space clustering, achieves comparable final scores only in sce-2 and sce-2m. It is attributed to the fact that clustering is most advantageous when units are dispersed and concentrated firepower is needed, whereas sce-2 and sce-2m begin with already concentrated formations. RL-Script, lacking both components, exhibits unstable training and significantly lower final scores, especially in sce-3m. In contrast, the full HRL-IM/CBS consistently outperforms every ablated variant in each scenario, confirming that each heuristic component contributes non-trivially to overall performance.

\begin{table*}[!ht]
\centering
\caption{Comparison of win rate and final score for different algorithms}
\label{tab:comparison}
\renewcommand{\arraystretch}{1.1}
\resizebox{0.96\textwidth}{!}{
\begin{tabular}{l|l|c|c|c|c|c|c}
\hline
 Metric & Algorithm & sce1 & sce1m & sce2 & sce2m & sce3 & sce3m \\ 
\hline
\multirow{7}{*}{Train Win Rate} & HRL-IM/CBS & \textbf{0.997 $\pm$ 0.005} & 0.980 $\pm$ 0.020 & \textbf{1.000 $\pm$ 0.000} & 0.997 $\pm$ 0.005 & 0.983 $\pm$ 0.015 & 0.898 $\pm$ 0.078 \\ 
 & MAPPO & 0.990 $\pm$ 0.012 & \textbf{0.998 $\pm$ 0.002} & \textbf{1.000 $\pm$ 0.000} & \textbf{1.000 $\pm$ 0.000} & \textbf{0.984 $\pm$ 0.010} & \textbf{1.000 $\pm$ 0.000} \\ 
 & QMIX & 0.925 $\pm$ 0.022 & 0.934 $\pm$ 0.021 & 0.992 $\pm$ 0.008 & 0.983 $\pm$ 0.005 & 0.973 $\pm$ 0.007 & 0.979 $\pm$ 0.001 \\ 
 & COMA & 0.911 $\pm$ 0.009 & 0.870 $\pm$ 0.014 & 0.994 $\pm$ 0.006 & 0.995 $\pm$ 0.005 & 0.976 $\pm$ 0.021 & 0.982 $\pm$ 0.011 \\ 
 & IQL & 0.904 $\pm$ 0.007 & 0.923 $\pm$ 0.007 & 0.996 $\pm$ 0.001 & 0.994 $\pm$ 0.001 & 0.946 $\pm$ 0.003 & 0.970 $\pm$ 0.002 \\ 
 & QTRAN & 0.895 $\pm$ 0.028 & 0.937 $\pm$ 0.010 & 0.993 $\pm$ 0.005 & 0.992 $\pm$ 0.006 & 0.948 $\pm$ 0.017 & 0.983 $\pm$ 0.003 \\ 
 & VDN & 0.920 $\pm$ 0.007 & 0.970 $\pm$ 0.001 & 0.993 $\pm$ 0.000 & 0.993 $\pm$ 0.002 & 0.954 $\pm$ 0.019 & 0.959 $\pm$ 0.003 \\ 
\hline
\multirow{7}{*}{Test Win Rate} & HRL-IM/CBS & 0.999 $\pm$ 0.003 & 0.985 $\pm$ 0.015 & 0.999 $\pm$ 0.003 & \textbf{1.000 $\pm$ 0.000} & 0.995 $\pm$ 0.005 & 0.925 $\pm$ 0.015 \\ 
 & MAPPO & \textbf{1.000 $\pm$ 0.000} & \textbf{1.000 $\pm$ 0.000} & \textbf{1.000 $\pm$ 0.000} & \textbf{1.000 $\pm$ 0.000} & \textbf{1.000 $\pm$ 0.000} & \textbf{1.000 $\pm$ 0.000} \\ 
 & QMIX & 0.988 $\pm$ 0.005 & 0.985 $\pm$ 0.015 & 0.994 $\pm$ 0.006 & \textbf{1.000 $\pm$ 0.000} & 0.952 $\pm$ 0.042 & 0.997 $\pm$ 0.003 \\ 
 & COMA & 0.887 $\pm$ 0.010 & 0.876 $\pm$ 0.024 & \textbf{1.000 $\pm$ 0.000} & \textbf{1.000 $\pm$ 0.000} & 0.975 $\pm$ 0.020 & 0.989 $\pm$ 0.011 \\ 
 & IQL & 0.967 $\pm$ 0.047 & \textbf{1.000 $\pm$ 0.000} & \textbf{1.000 $\pm$ 0.000} & \textbf{1.000 $\pm$ 0.000} & 0.950 $\pm$ 0.050 & \textbf{1.000 $\pm$ 0.000} \\ 
 & QTRAN & \textbf{1.000 $\pm$ 0.000} & 0.994 $\pm$ 0.006 & \textbf{1.000 $\pm$ 0.000} & \textbf{1.000 $\pm$ 0.000} & 0.985 $\pm$ 0.015 & 0.997 $\pm$ 0.003 \\ 
 & VDN & 0.988 $\pm$ 0.012 & 0.997 $\pm$ 0.003 & \textbf{1.000 $\pm$ 0.000} & 0.997 $\pm$ 0.003 & 0.994 $\pm$ 0.006 & \textbf{1.000 $\pm$ 0.000} \\ 
\hline
\multirow{7}{*}{Train Final Score} & HRL-IM/CBS & \textbf{47.847 $\pm$ 1.233} & 24.513 $\pm$ 0.361 & \textbf{77.246 $\pm$ 0.538} & 51.773 $\pm$ 4.190 & 46.183 $\pm$ 5.288 & 44.794 $\pm$ 5.823 \\ 
 & MAPPO & 20.777 $\pm$ 0.857 & \textbf{35.849 $\pm$ 2.840} & 65.389 $\pm$ 0.987 & \textbf{61.892 $\pm$ 0.663} & 47.739 $\pm$ 4.606 & \textbf{75.329 $\pm$ 1.970} \\ 
 & QMIX & 13.765 $\pm$ 1.958 & 21.055 $\pm$ 2.081 & 39.843 $\pm$ 0.607 & 33.401 $\pm$ 0.187 & 40.818 $\pm$ 2.478 & 42.576 $\pm$ 2.863 \\ 
 & COMA & 16.356 $\pm$ 1.248 & 18.107 $\pm$ 1.345 & 37.538 $\pm$ 2.005 & 40.988 $\pm$ 0.181 & \textbf{51.514 $\pm$ 2.466} & 57.033 $\pm$ 1.749 \\ 
 & IQL & 11.782 $\pm$ 1.092 & 19.104 $\pm$ 1.180 & 47.096 $\pm$ 0.135 & 42.589 $\pm$ 4.474 & 33.878 $\pm$ 0.334 & 40.839 $\pm$ 1.371 \\ 
 & QTRAN & 14.389 $\pm$ 3.044 & 19.946 $\pm$ 2.129 & 29.937 $\pm$ 1.329 & 38.835 $\pm$ 0.065 & 38.447 $\pm$ 5.164 & 45.814 $\pm$ 0.387 \\ 
 & VDN & 13.110 $\pm$ 0.514 & 23.504 $\pm$ 0.581 & 31.464 $\pm$ 0.308 & 32.062 $\pm$ 1.718 & 35.546 $\pm$ 1.288 & 42.816 $\pm$ 0.461 \\ 
\hline
\multirow{7}{*}{Test Final Score} & HRL-IM/CBS & \textbf{48.959 $\pm$ 2.444} & 25.697 $\pm$ 0.356 & \textbf{78.109 $\pm$ 0.861} & 52.409 $\pm$ 4.625 & 48.332 $\pm$ 4.954 & 47.158 $\pm$ 2.665 \\ 
 & MAPPO & 22.250 $\pm$ 1.912 & \textbf{37.406 $\pm$ 3.712} & 64.500 $\pm$ 1.723 & \textbf{62.500 $\pm$ 1.794} & 50.281 $\pm$ 5.246 & \textbf{74.406 $\pm$ 0.840} \\ 
 & QMIX & 17.989 $\pm$ 1.347 & 25.797 $\pm$ 3.763 & 43.952 $\pm$ 0.711 & 31.710 $\pm$ 4.924 & 40.295 $\pm$ 6.808 & 56.587 $\pm$ 1.299 \\ 
 & COMA & 14.648 $\pm$ 1.316 & 19.137 $\pm$ 3.337 & 38.494 $\pm$ 2.174 & 42.572 $\pm$ 0.927 & \textbf{55.266 $\pm$ 0.586} & 55.375 $\pm$ 5.761 \\ 
 & IQL & 21.736 $\pm$ 2.742 & 26.666 $\pm$ 9.069 & 56.307 $\pm$ 0.898 & 46.794 $\pm$ 8.827 & 46.604 $\pm$ 0.543 & 48.082 $\pm$ 1.420 \\ 
 & QTRAN & 20.461 $\pm$ 3.963 & 27.993 $\pm$ 0.993 & 33.299 $\pm$ 0.787 & 43.598 $\pm$ 1.527 & 44.532 $\pm$ 2.851 & 57.920 $\pm$ 7.448 \\ 
 & VDN & 17.532 $\pm$ 2.837 & 29.794 $\pm$ 3.504 & 27.071 $\pm$ 2.597 & 32.677 $\pm$ 3.083 & 53.321 $\pm$ 6.969 & 47.620 $\pm$ 9.024 \\ 
\hline
\end{tabular}
}
\end{table*}

\subsection{Comparison Results}
Figure~\ref{fig:comparison} illustrates the performance of HRL-IM/CBS and other algorithms across all scenarios during both training and testing stages. The detailed results of win rate and final score, including mean and standard deviation, are presented in Table~\ref{tab:comparison}. In all scenarios, the metric win rate typically increases with the training process and eventually approaches 1, and the final score also improves to some extent. 

Quantitative results show that, except for COMA leading on sce-3, MAPPO achieves overwhelming superiority over all other baselines and delivers remarkable performance. In sce-1 and sce-2, the proposed HRL-IM/CBS significantly surpasses MAPPO, maintaining a high win rate while obtaining higher final scores, demonstrating the critical role of cluster-based scripts in guiding the algorithm toward high-quality solutions. In sce-1m and sce-2m, HRL-IM/CBS achieves the second-highest final score, marginally behind MAPPO. In sce-3 and sce-3m, the win rate of HRL-IM/CBS exhibits fluctuations during the training phase fluctuates. Nevertheless, it attains a final score comparable to those of other algorithms.

The performance differences of HRL-IM/CBS across different scenarios are primarily attributed to the variations in unit distribution and scenario complexity. Between 4v4 scale scenarios (sce-1/1m and sce-2/2m), the unit distribution in sce-1 and sce-1m is more concentrated compared to sce-2 and sce-2m. This leads to a lower win rate and final score in the early training stage. For 8v8 scale scenarios, the formation size and unit distribution in sce-3 and sce-3m are more intricate, resulting in a relatively lower win rate and final score in the early training stage compared to other scenarios. When comparing the original scenarios with their mirror scenarios, there are notable differences in final score, especially in sce-2 and sce-2m. This indicates significant disparities in the strength of both sides between the original and mirror scenarios. Despite these differences, the algorithm remains effective in both scenarios.

It should also be highlighted that a minor performance gap exists between training and testing phases. Even with {$\varepsilon$} reduced to 0.05 in late training, low exploration rewards accumulate due to the multi-table architecture and long-term decisions. In testing, the {$\varepsilon$}\text{-greedy} mechanism is deactivated to select higher-reward actions.

\subsection{Effectiveness and Mechanism Analysis}

\begin{figure*}[!th]
  \centering
  \begin{minipage}{1.0\textwidth}
    \centering
    \includegraphics[width=\columnwidth]{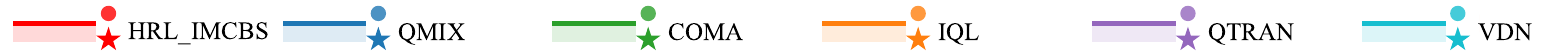}
  \end{minipage}\hfill

  \medskip
  
  \begin{minipage}{0.32\textwidth}
    \centering
    \includegraphics[width=\columnwidth]{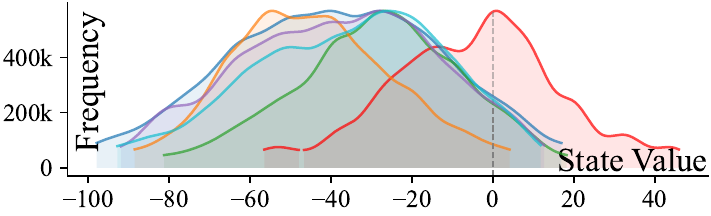}
    \small{(a) sce-1}
  \end{minipage}%
  \begin{minipage}{0.32\textwidth}
    \centering
    \includegraphics[width=\columnwidth]{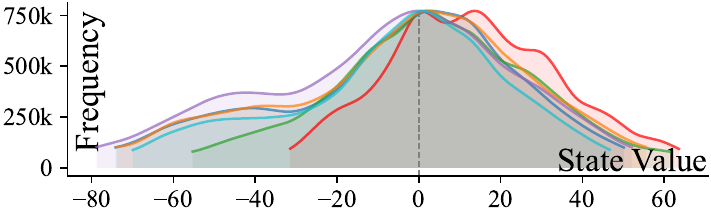}
    \small{(b) sce-2}
  \end{minipage}%
  \begin{minipage}{0.32\textwidth}
    \centering
    \includegraphics[width=\columnwidth]{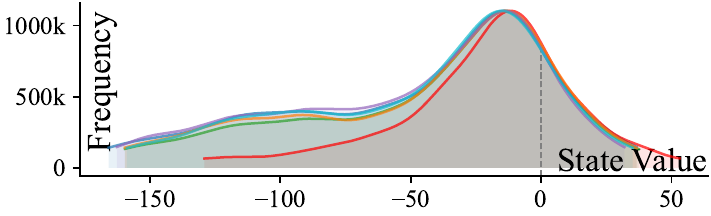}
    \small{(c) sce-3}
  \end{minipage}%

  \medskip

  \begin{minipage}{0.32\textwidth}
    \centering
    \includegraphics[width=\columnwidth]{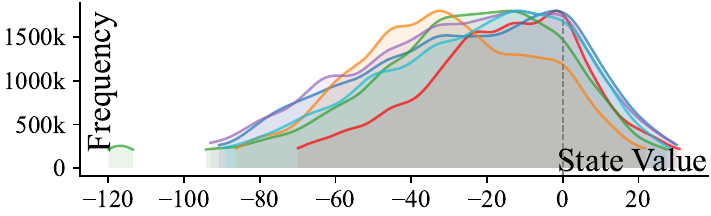}
    \small{(d) sce-1m}
  \end{minipage}%
  \begin{minipage}{0.32\textwidth}
    \centering
    \includegraphics[width=\columnwidth]{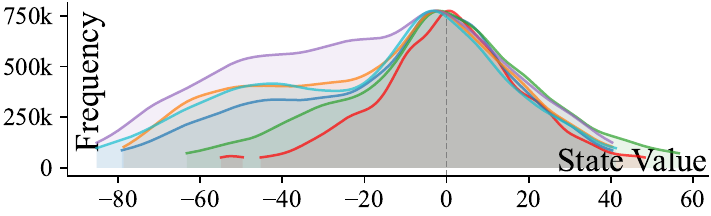}
    \small{(e) sce-2m}
  \end{minipage}%
  \begin{minipage}{0.32\textwidth}
    \centering
    \includegraphics[width=\columnwidth]{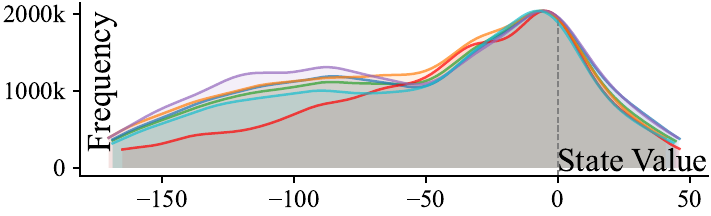}
    \small{(f) sce-3m}
  \end{minipage}%

  \caption{State-sampling frequency weighted density ($95.4\%$ HDI, $\approx\pm 2\sigma$) for different algorithms across all scenarios.}
  \label{fig:state sampling frequency}
\end{figure*}

Comprehensive effectiveness and mechanism analyses are conducted to clarify the advantages of HRL-IM/CBS, presenting insights from three perspectives including sample efficiency, training efficiency and interpretability, along with attribution analyses focusing on proposed components.

\subsubsection{Sample Efficiency}
To clearly demonstrate the sample-efficient advantages, the distributional characteristics of the states sampled by algorithms were analyzed from both state-space coverage and statistical-numerical perspective. The sampling state density bands have been estimated with a kernel density estimator that employs a Gaussian kernel of bandwidth 0.15 (state-value units). Each state is weighted by its total visitation count by the given algorithm, so the resulting density is proportional to the actual sampling frequency. Each shaded band is set to the $95.4\%$ highest-density region, corresponding to the $\pm 2\sigma$ interval of a normal distribution and encompassing the bulk of the samples.

Figure~\ref{fig:state sampling frequency} shows that in sce-1 and sce-2, the density bands of states sampled by HRL-IM/CBS are visibly skewed toward higher values. In the other scenarios, low-value regions of density bands converge toward high-value areas, indicating that the algorithm still avoids exploring unfavorable states. This pattern fundamentally demonstrates the algorithm’s ability to actively favor advantageous states and avoid disadvantageous ones, which directly accelerates training and improves final performance. The detailed distribution of sampled states over state value landscape are shown in Figure~A.1 in Appendix A, which reveals the rugged nature of the state-value landscape.

The advantages demonstrated by these results are primarily attributed to influence map hashing achieving accurate state space division and efficient collaboration based on cluster-based scripts. These two factors respectively avoid frequent sampling of inefficient states from the perspectives of accurate state partitioning and efficient exploration behavior, thereby enhancing the sample-efficient advantage of HRL-IM/CBS.

\subsubsection{Training Efficiency}

\begin{figure*}[!th]
    \centering
    \begin{minipage}{0.32\textwidth}
        \centering
        \includegraphics[width=\columnwidth]{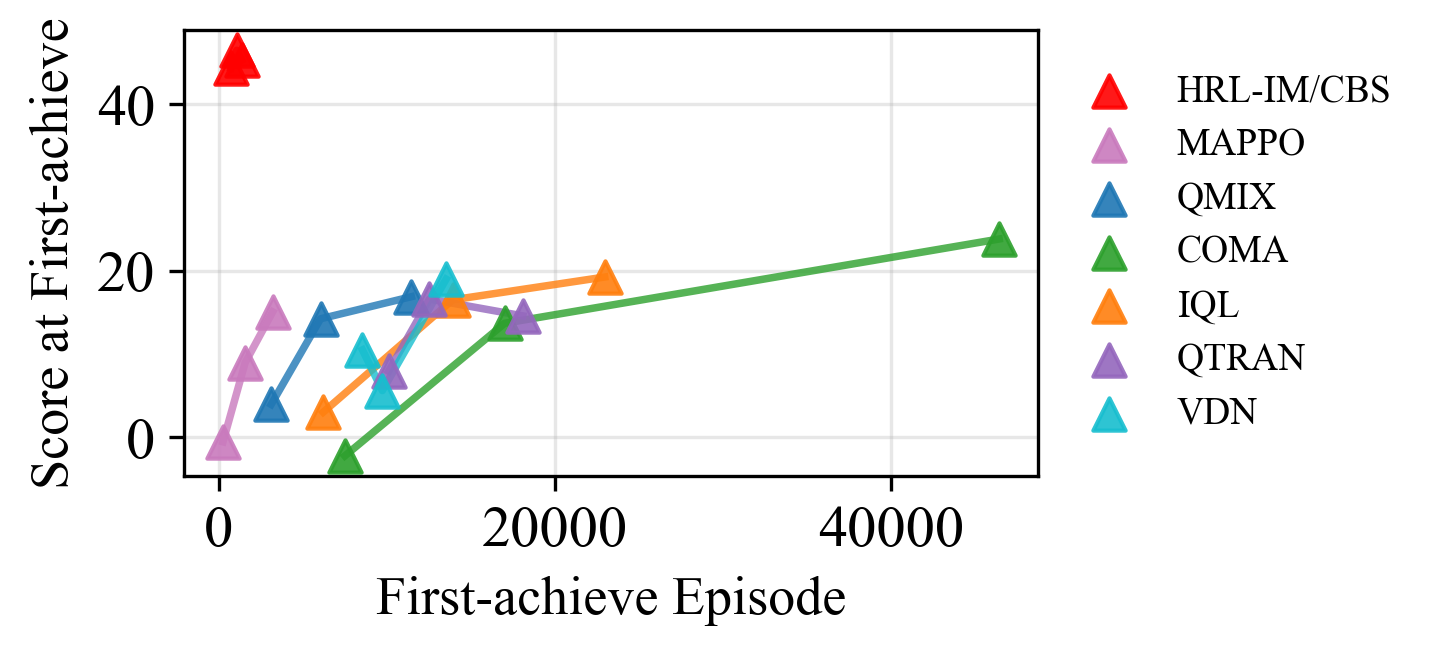}
        \small{(a) sce-1}
    \end{minipage}
    \begin{minipage}{0.32\textwidth}
        \centering
        \includegraphics[width=\columnwidth]{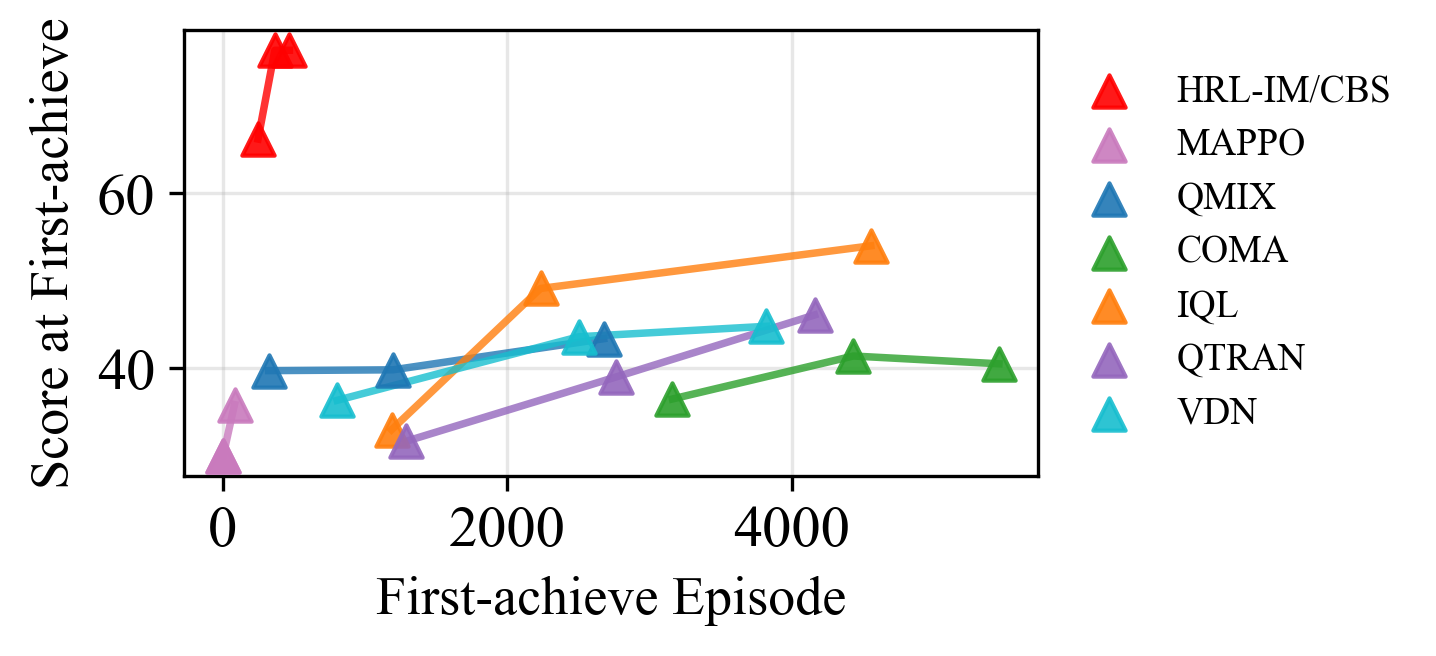}
        \small{(b) sce-2}
    \end{minipage}
    \begin{minipage}{0.32\textwidth}
        \centering
        \includegraphics[width=\columnwidth]{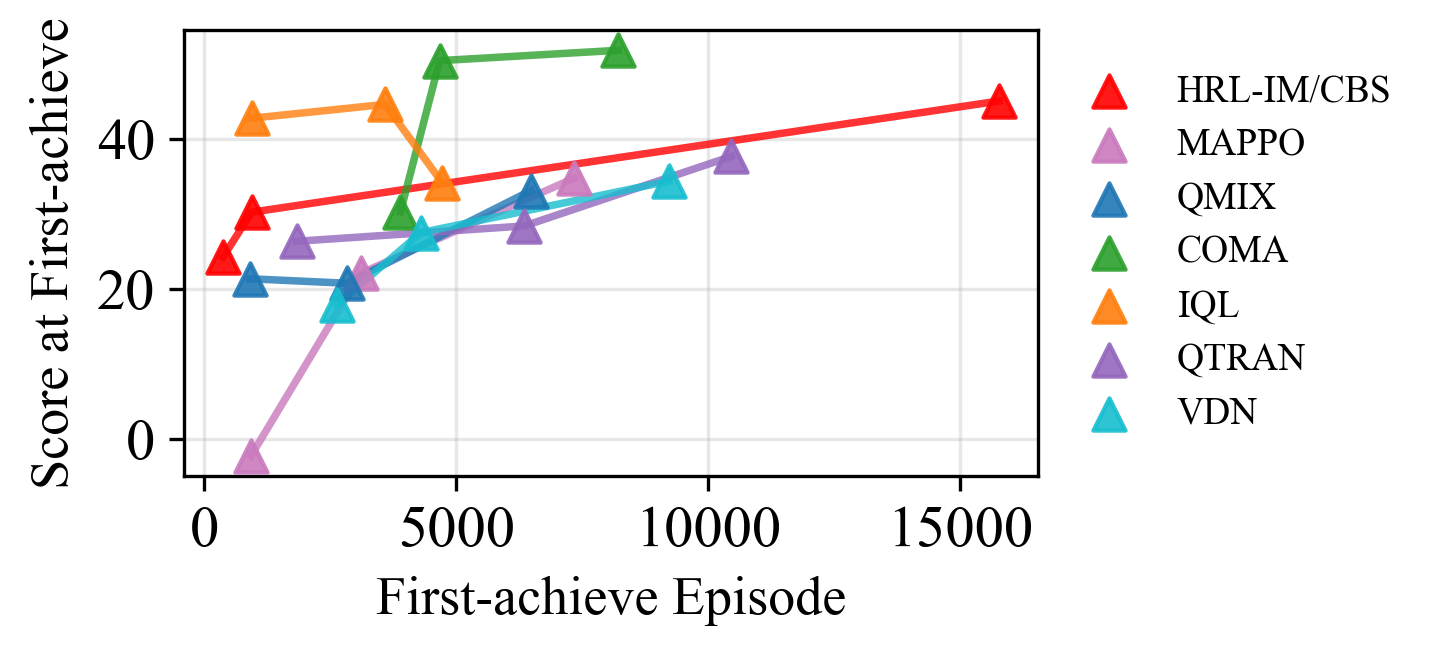}
        \small{(c) sce-3}
    \end{minipage}

    \begin{minipage}{0.32\textwidth}
        \centering
        \includegraphics[width=\columnwidth]{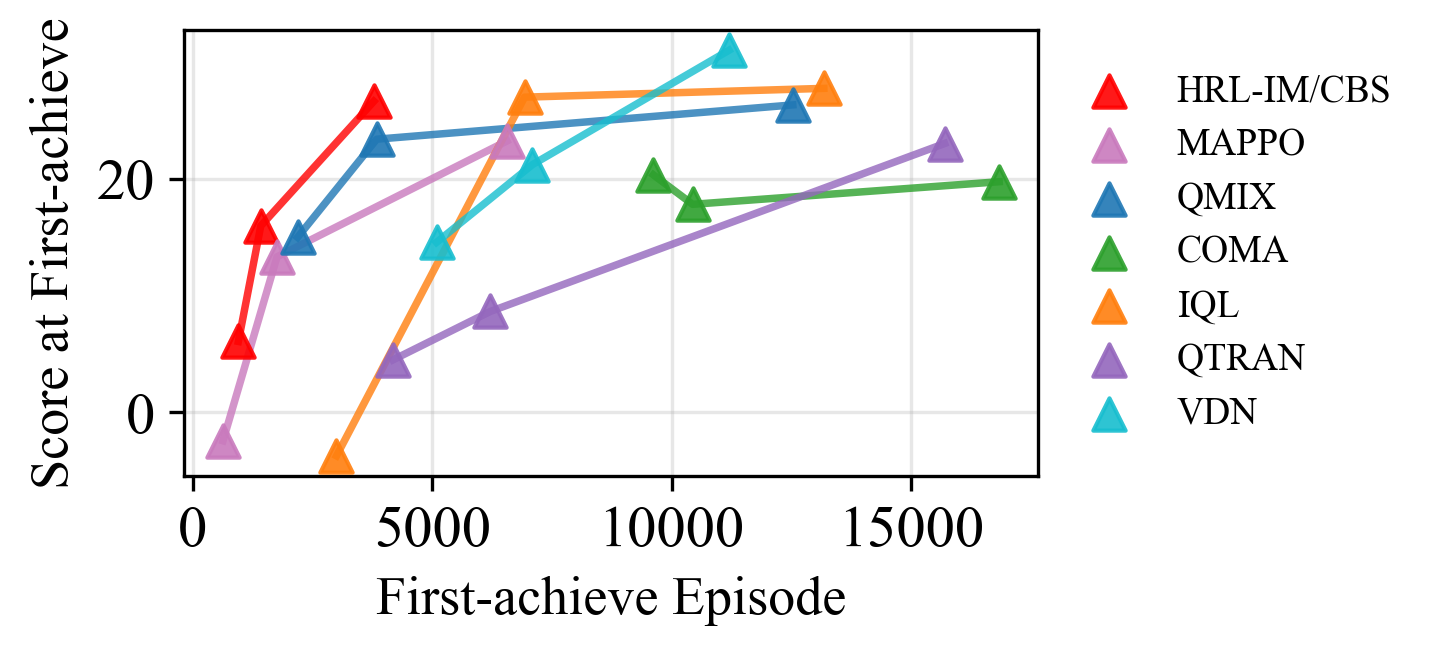}
        \small{(d) sce-1m}
    \end{minipage}
    \begin{minipage}{0.32\textwidth}
        \centering
        \includegraphics[width=\columnwidth]{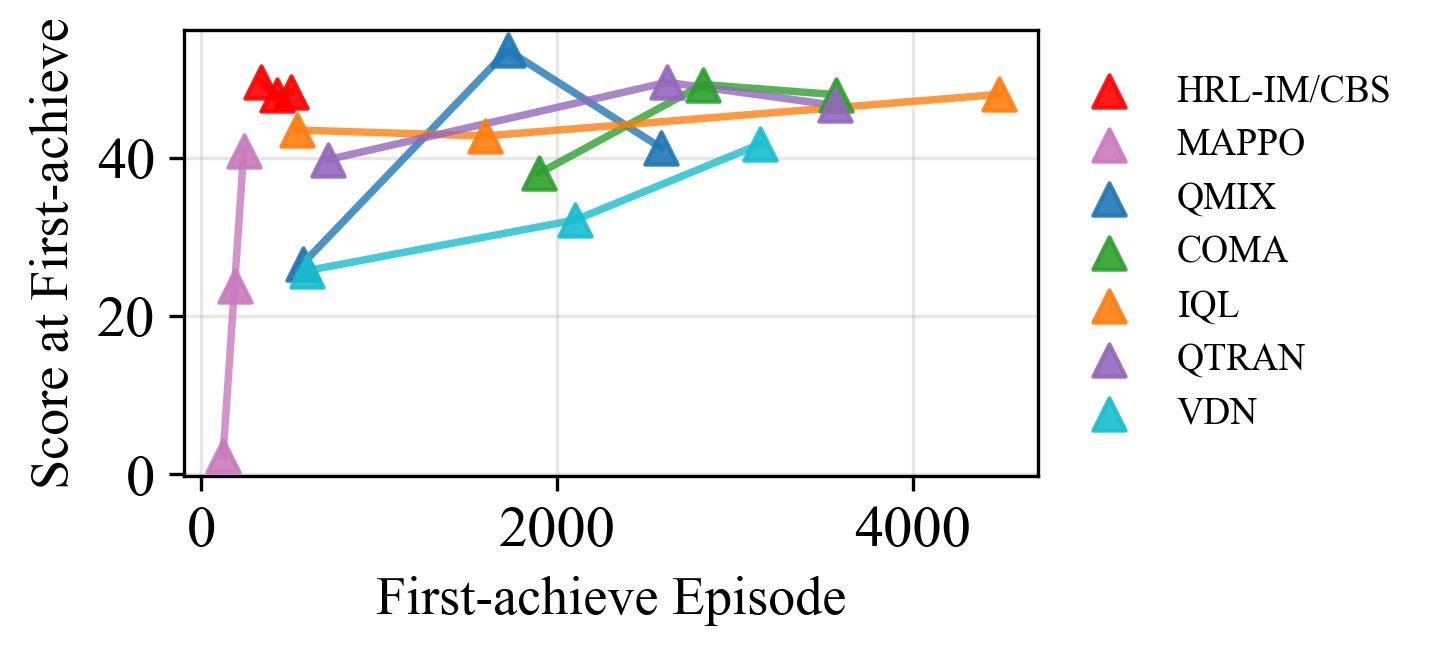}
        \small{(e) sce-2m}
    \end{minipage}
    \begin{minipage}{0.32\textwidth}
        \centering
        \includegraphics[width=\columnwidth]{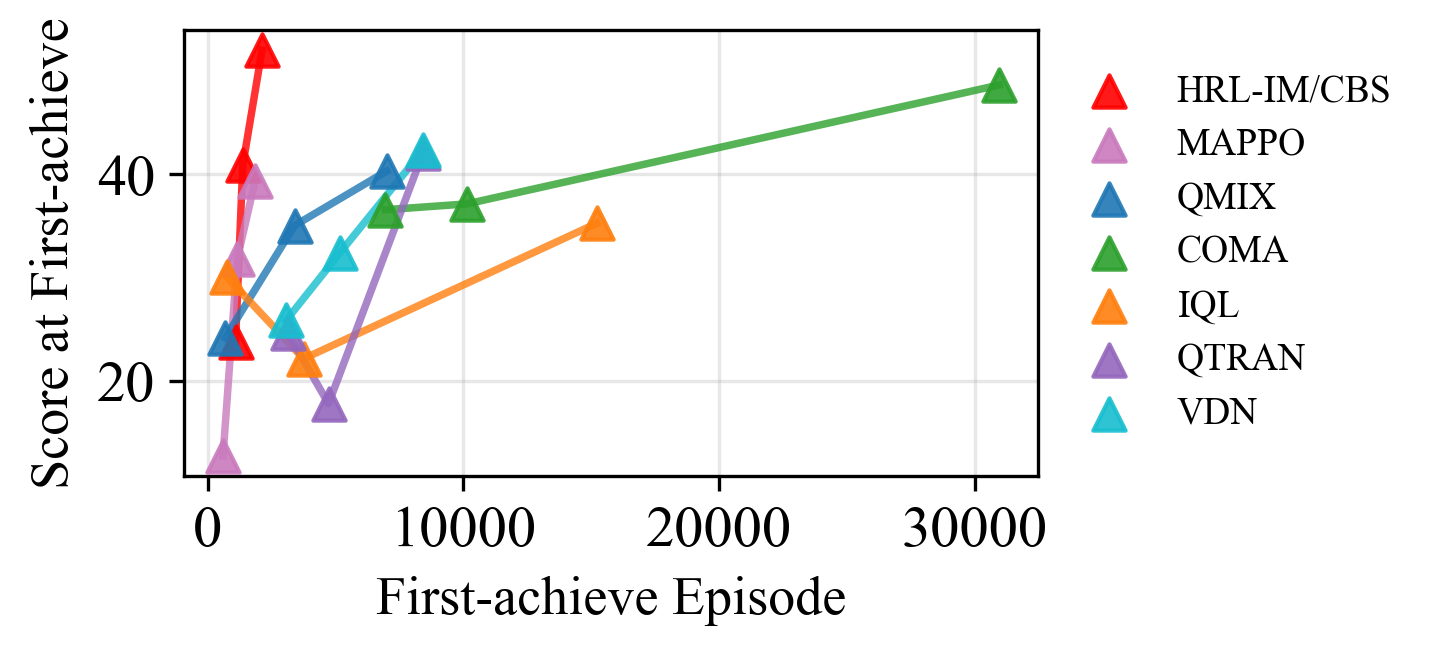}
        \small{(f) sce-3m}
    \end{minipage}
    
\caption{First-achievement trajectories of 50\%/75\%/95\% test win rates for different algorithms in all scenarios.}
\label{fig:threshold-achievement}
\end{figure*}

To demonstrate the training efficiency of the proposed method, the training episodes required to first achieve test win rates of 50\%, 75\% and 95\% were recorded, along with their corresponding final scores to measure the speed of finding high-quality solutions. The vertical axis represents the number of training episodes, while the horizontal axis shows the corresponding final scores obtained in evaluation. Algorithm trajectories with superior sample efficiency are located in the upper-left region, representing high returns achieved within fewer episodes.

Figure~\ref{fig:threshold-achievement} shows that in sce-1 and sce-2, HRL-IM/CBS delivers markedly superior performance. The training curves remain stable and the final scores significantly exceed those of the baselines, while the first-achievement trajectories lie in the upper-left region, confirming that high-quality solutions are discovered early in training. Across the three mirrored scenarios, HRL-IM/CBS maintains a consistent sample efficiency advantage. In sce-1m, it reaches the 95\% win rate with fewer episodes and a higher final score than MAPPO. In sce-2m, its episode count is marginally higher than MAPPO’s, yet the obtained score is substantially superior. the slightly elevated score at the 50\% win rate point reflects a transient local optimum encountered early in training. In sce-3m, the episode counts required to attain the target win rates are comparable to MAPPO, but HRL-IM/CBS again achieves markedly higher scores.

\begin{figure*}[!th]
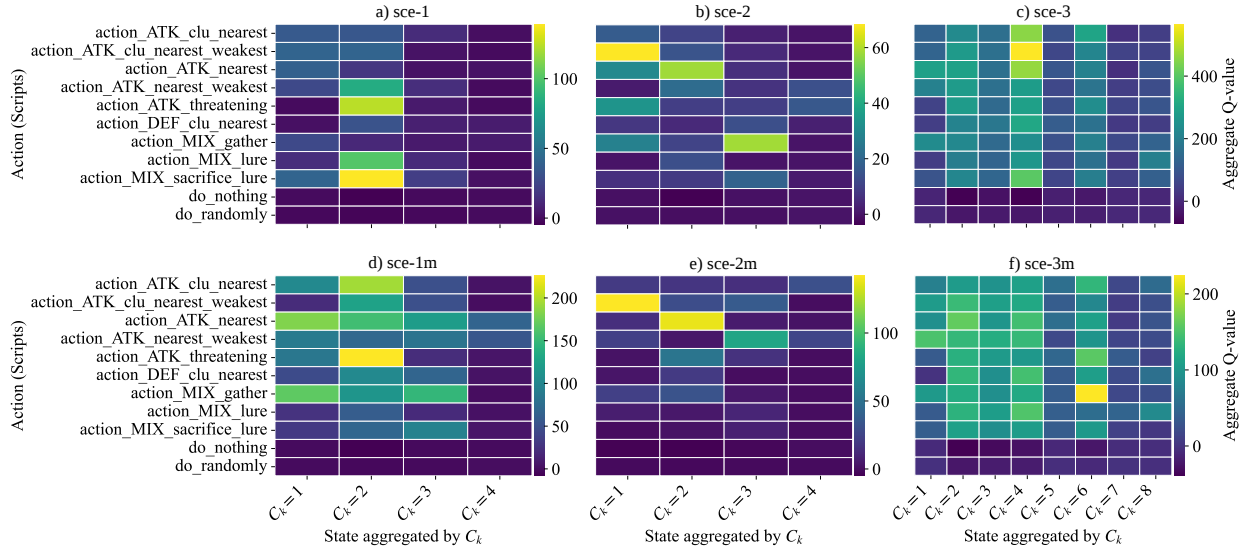

    \centering
    \begin{minipage}{\textwidth}
        \centering
        \includegraphics[width=\columnwidth]{strategy_1.pdf}
    \end{minipage}
    
    \begin{minipage}{\textwidth}
        \centering
        \includegraphics[width=\columnwidth]{strategy_2.pdf}
    \end{minipage}
    \caption{Q-value heatmaps under aggregated states by combat space clustering of HRL-IM/CBS across all scenarios.}
    \label{fig:strategy}
\end{figure*}

In sce-3, HRL-IM/CBS attains a final score statistically indistinguishable from the baselines, yet exhibits pronounced win rate volatility during early training and surpasses the 95\% threshold after approximately 15000 episodes, indicating potential enhancement in high-complexity scenarios. Quantitative results confirm that HRL-IM/CBS is able to locate high-quality solutions, yet its robustness remains limited, meaning that small perturbations of the input space can induce large output variations. Apart from the inherent stochasticity and complexity of the RTS game, the degradation stems primarily from the reduced accuracy of influence map hashing when confronted with a vast and intricate state space, thereby degrading action selection consistency.

The advantages demonstrated by these results are primarily attributed to efficient model updates enabled by substantial state space compression through influence map hashing, high-quality sampling driven by cluster-based scripts from early training stages, and efficient sample utilization in sparse reward environments facilitated by hierarchical architecture and reward allocation mechanisms. These components collectively enhance training efficiency from different perspectives.

\subsubsection{Interpretability}
The interpretability of the proposed HRL-IM/CBS stems from the refined Q-tables, which explicitly store the expected return for each state–action pair. To visualize this property, Q-value heatmaps are employed, which provide an immediate and intuitive interpretation of the learned policy. Owing to the hierarchical multi-Q-table structure, a single upper-level state may correspond to multiple clustering schemes, each with distinct actions and Q-values. This results in a direct heatmap of raw states is infeasible. Fortunately, combat space clustering enables effortless analysis of the Q-values associated with scripts under each clustering scheme, thereby providing clear, interpretable insights.

To demonstrate the advantages of HRL-IM/CBS in terms of interpretability, Figure~\ref{fig:strategy} presents the Q-value heatmaps produced by HRL-IM/CBS under aggregated states obtained through combat space clustering. The vertical axis lists all cluster-based scripts and the horizontal axis indexes aggregated states by $C_k$, with each column integrating Q-values across all partitions that cluster the raw state into $C_k$ clusters.

In Figure~\ref{fig:strategy}, the Q-value heatmaps across scenarios exhibit distinct tactical preferences. In sce-1 and sce-1m, concentrating fire on the most threatening enemy unit under $C_k = 2$ proves highly effective, while in sce-1 the sacrifice-lure tactic also performs well, aligning perfectly with the scenarios' deceptive unit placements. Moreover, in both scenarios, combat space clustering consistently produces patently unreasonable groupings at $C_k = 4$, accompanied by markedly lower Q-values.

Scenarios sce-2 and sce-2m exhibit different deployments yet demand similar tactical preferences. In sce-2, self units encircle the enemy and sequential focused fire is optimal, as evidenced by the lowest Q-value assigned to the script \texttt{\small action\_ATK\_nearest\_weakest} that disperses attacks. In sce-2m, self units are clustered and the enemy is dispersed, resulting in the highest Q-value for the script \texttt{\small action\_ATK\_clu\_nearest\_weakest} which directs all units to simultaneously attack the globally nearest weakest target.

Scenarios sce-3 and sce-3m present significantly higher complexity, and the corresponding Q-value heatmaps reveal richly diversified tactical preferences. The observed heterogeneity across scripts and clusters demonstrates that HRL-IM/CBS flexibly acquires a diverse repertoire of locally conditioned tactics, thereby precluding convergence to a single, globally uniform policy. Additionally, the similarity in unit distribution across the two scenarios yields congruent patterns across several clustering schemes. Figure~\ref{fig:strategy} c) and f) indicate that the clustering schemes corresponding to $C_k = 5, 7, 8$ are not reasonable, which is determined by the characteristics of the unit distribution. For $C_k = 1$, the script \texttt{\small action\_DEF\_clu\_nearest\_weakest} is evidently inappropriate, as the interlaced formation of units of both sides makes collective retreat extremely unfavorable. Moreover, across all scenarios, the scripts \texttt{\small do\_nothing} and \texttt{\small do\_randomly} consistently yield the lowest Q-values, as expected.

The aforementioned results provide interpretable analysis of tactical preferences, with this advantage attributed to the focus of combat space clustering on local coordination and generalization of combat situations. Additionally, the multi-Q-table model fundamentally preserves rewards for state-action pairs, providing a foundation for clustering-based analysis.

\vspace{1em}
\section{Conclusion}
\label{Conclusion}
In this paper, a HRL-IM/CBS framework is proposed combined with influence map and script methods for RTS micro. To effectively reduce the size of the state and action space, an influence map hashing mechanism is designed to express the high-dimensional combat space information and cluster-based script is introduced to ensure the flexible control of global team or local units. With a hierarchical Q-learning architecture combined with combat space clustering and reward allocation mechanism, the tactical collaboration can be decomposed into upper and lower level tasks. For the upper level, there is one Q-table representing team tactics at the spatial clustering level, while for the lower level, numerous Q-tables represent the regulations of unit actions at the combat level. With utilizing only classical Q-learning for training from scratch, a lightweight but interpretable model has been created, which has been verified by experimental analysis on the fair scenarios established in this study. By directly deploying the trained model on the original map and its mirror map, the proposed method has demonstrated effectiveness and wide suitability on different scenarios.

In future work, the application of machine-learning techniques for automatic script extraction is planned to be investigated instead of relying on manual design. Furthermore, exploration will be conducted on how the trained model can serve as a forward knowledge model to support advanced algorithms such as evolutionary computation, facilitating the generation of high-quality solutions and providing guidance for evolution. In parallel, real-time clustering of game-state data streams and the corresponding state-distance metrics will be studied, so as to pave the way for defining neighborhood relationships among RTS game solution sequences. Through these efforts, it is anticipated that this trustworthy, domain-knowledge-driven model will play a greater role in both AI research and real-world multi-agent collaborative scheduling problems that share essential similarities with RTS micromanagement.

\bibliographystyle{IEEEtran}
\bibliography{ref}

\clearpage
\appendix
\setcounter{equation}{0}
\setcounter{figure}{0}
\setcounter{table}{0}
\renewcommand{\theequation}{A.\arabic{equation}}
\renewcommand{\thefigure}{A.\arabic{figure}}
\renewcommand{\thetable}{A.\arabic{table}}
\renewcommand{\theHequation}{A.\arabic{equation}}
\renewcommand{\theHfigure}{A.\arabic{figure}}
\renewcommand{\theHtable}{A.\arabic{table}}
The supplemental file presents the comprehensive experimental results of ablation study and the visualization of the distribution of sampled sates.

\section{Supplementary Material}
\label{secA}

\subsection{Ablation Study}

\begin{table*}[!ht]
\centering
\caption{Component ablation study of HRL-IM/CBS across all scenarios}
\label{tab:ablation}
\resizebox{0.96\textwidth}{!}{
\begin{tabular}{l|l|c|c|c|c|c|c}
\hline
 Metric & Algorithm & sce1 & sce1m & sce2 & sce2m & sce3 & sce3m \\ 
\hline
\multirow{4}{*}{Train Win Rate} & HRL-IM/CBS & \textbf{0.997 $\pm$ 0.005} & \textbf{0.980 $\pm$ 0.020} & \textbf{1.000 $\pm$ 0.000} & \textbf{0.997 $\pm$ 0.005} & \textbf{0.983 $\pm$ 0.015} & \textbf{0.898 $\pm$ 0.078} \\ 
 & HRL-CBS & 0.990 $\pm$ 0.008 & 0.842 $\pm$ 0.113 & 0.995 $\pm$ 0.005 & 0.997 $\pm$ 0.007 & 0.852 $\pm$ 0.075 & 0.427 $\pm$ 0.232 \\ 
 & RL-IM/Script & 0.892 $\pm$ 0.090 & 0.578 $\pm$ 0.204 & \textbf{1.000 $\pm$ 0.000} & 0.927 $\pm$ 0.054 & 0.792 $\pm$ 0.128 & 0.204 $\pm$ 0.082 \\ 
 & RL-Script & 0.683 $\pm$ 0.205 & 0.614 $\pm$ 0.230 & 0.921 $\pm$ 0.069 & 0.485 $\pm$ 0.475 & 0.871 $\pm$ 0.096 & 0.338 $\pm$ 0.338 \\ 
\hline
\multirow{4}{*}{Test Win Rate} & HRL-IM/CBS & \textbf{0.999 $\pm$ 0.003} & \textbf{0.985 $\pm$ 0.015} & 0.999 $\pm$ 0.003 & \textbf{1.000 $\pm$ 0.000} & \textbf{0.995 $\pm$ 0.005} & \textbf{0.925 $\pm$ 0.015} \\ 
 & HRL-CBS & 0.990 $\pm$ 0.014 & 0.831 $\pm$ 0.093 & \textbf{1.000 $\pm$ 0.000} & \textbf{1.000 $\pm$ 0.000} & 0.890 $\pm$ 0.070 & 0.485 $\pm$ 0.207 \\ 
 & RL-IM/Script & 0.903 $\pm$ 0.071 & 0.573 $\pm$ 0.224 & 0.988 $\pm$ 0.016 & 0.936 $\pm$ 0.064 & 0.857 $\pm$ 0.093 & 0.221 $\pm$ 0.106 \\ 
 & RL-Script & 0.727 $\pm$ 0.184 & 0.581 $\pm$ 0.213 & 0.920 $\pm$ 0.092 & 0.500 $\pm$ 0.500 & 0.845 $\pm$ 0.116 & 0.370 $\pm$ 0.370 \\ 
\hline
\multirow{4}{*}{Train Final Score} & HRL-IM/CBS & \textbf{47.847 $\pm$ 1.233} & \textbf{24.513 $\pm$ 0.361} & \textbf{77.246 $\pm$ 0.538} & \textbf{51.773 $\pm$ 4.190} & \textbf{46.183 $\pm$ 5.288} & \textbf{44.794 $\pm$ 5.823} \\ 
 & HRL-CBS & 43.231 $\pm$ 2.256 & 15.643 $\pm$ 5.022 & 59.965 $\pm$ 12.538 & 51.233 $\pm$ 3.369 & 37.285 $\pm$ 13.045 & -20.169 $\pm$ 21.769 \\ 
 & RL-IM/Script & 21.019 $\pm$ 23.114 & 2.273 $\pm$ 7.578 & 63.184 $\pm$ 17.537 & 44.274 $\pm$ 2.447 & 26.985 $\pm$ 18.872 & -17.499 $\pm$ 5.707 \\ 
 & RL-Script & 7.800 $\pm$ 18.638 & 5.352 $\pm$ 10.785 & 21.098 $\pm$ 2.879 & 11.071 $\pm$ 23.395 & 28.430 $\pm$ 14.378 & -29.963 $\pm$ 32.142 \\ 
\hline
\multirow{4}{*}{Test Final Score} & HRL-IM/CBS & \textbf{48.959 $\pm$ 2.444} & \textbf{25.697 $\pm$ 0.356} & \textbf{78.109 $\pm$ 0.861} & \textbf{52.409 $\pm$ 4.625} & \textbf{48.332 $\pm$ 4.954} & \textbf{47.158 $\pm$ 2.665} \\ 
 & HRL-CBS & 43.475 $\pm$ 2.362 & 15.048 $\pm$ 5.258 & 62.203 $\pm$ 12.724 & 52.238 $\pm$ 3.511 & 42.479 $\pm$ 12.881 & -15.163 $\pm$ 21.747 \\ 
 & RL-IM/Script & 19.151 $\pm$ 22.818 & 6.159 $\pm$ 9.462 & 64.259 $\pm$ 17.850 & 47.080 $\pm$ 2.350 & 32.777 $\pm$ 16.993 & -16.080 $\pm$ 7.368 \\ 
 & RL-Script & 11.569 $\pm$ 17.629 & 2.830 $\pm$ 12.140 & 22.288 $\pm$ 4.696 & 11.809 $\pm$ 24.589 & 28.092 $\pm$ 18.769 & -30.367 $\pm$ 35.178 \\ 
\hline
\end{tabular}
}
\end{table*}

Table~\ref{tab:ablation} shows the quantitative experimental results of different ablation variants in all scenarios, including win rates and final scores. The experimental results demonstrate that the complete HRL-IM/CBS outperforms all ablation variants, proving the unique contribution of each component. Notably, in sce-3m, all ablation variants fail to achieve satisfactory results, with win rates below 50\%.

\subsection{Distribution of Sampled States}

\begin{figure}[!ht]
  \centering
  \begin{minipage}{0.9\textwidth}
    \centering
    \includegraphics[width=\columnwidth]{legend.pdf}
  \end{minipage}\hfill

  \medskip
  
  \begin{minipage}{0.28\textwidth}
    \centering
    \includegraphics[width=\columnwidth]{MarineMicro_MvsM_4_state_kde_band_tight_single.pdf}
  \end{minipage}%
  \begin{minipage}{0.28\textwidth}
    \centering
    \includegraphics[width=\columnwidth]{MarineMicro_MvsM_4_dist_state_kde_band_tight_single.pdf}
  \end{minipage}%
  \begin{minipage}{0.28\textwidth}
    \centering
    \includegraphics[width=\columnwidth]{MarineMicro_MvsM_8_state_kde_band_tight_single.pdf}
  \end{minipage}%

  \begin{minipage}{0.28\textwidth}
    \centering
    \includegraphics[width=\columnwidth]{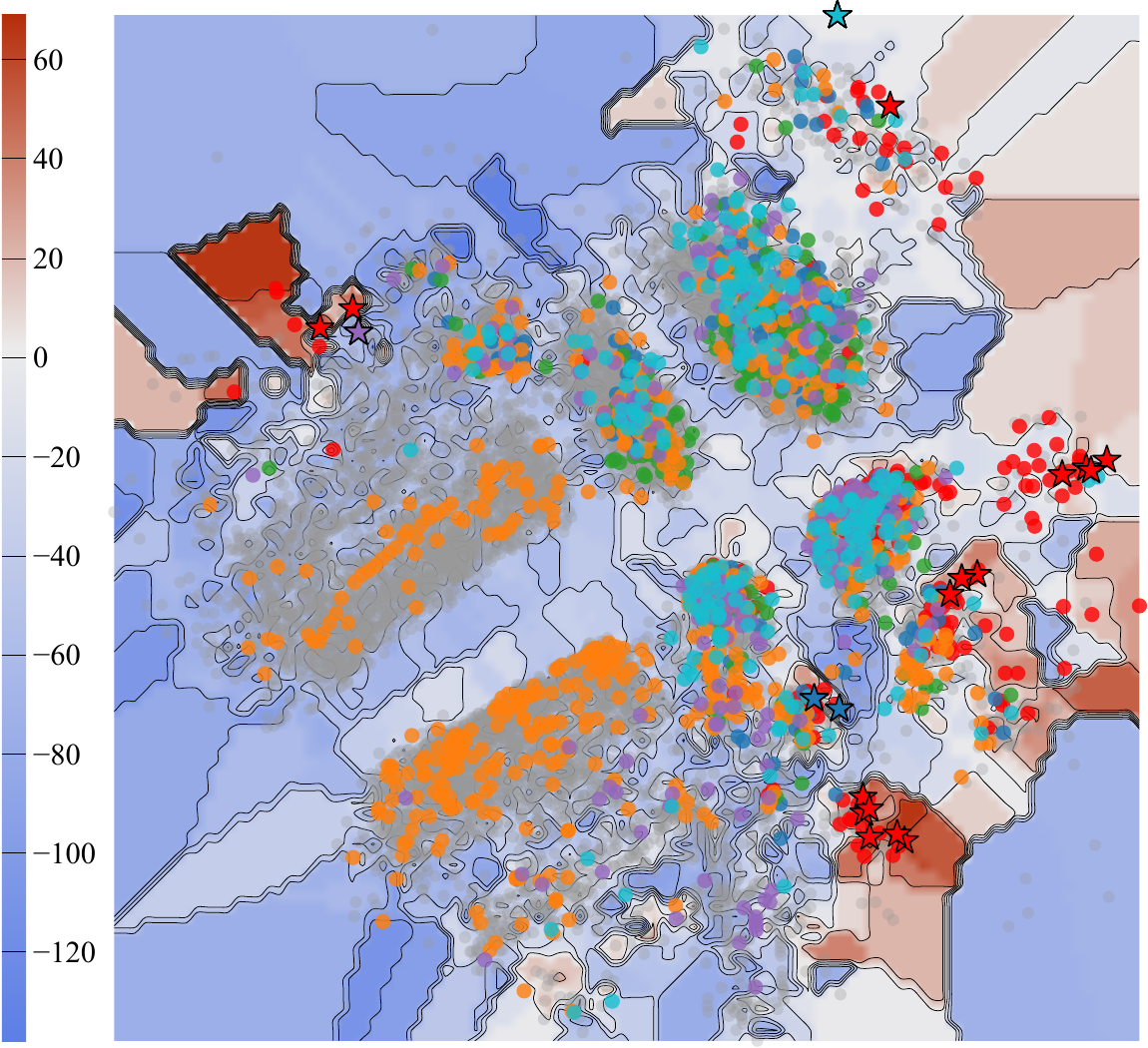}
     \small{(a) sce-1}
  \end{minipage}%
  \begin{minipage}{0.28\textwidth}
    \centering
    \includegraphics[width=\columnwidth]{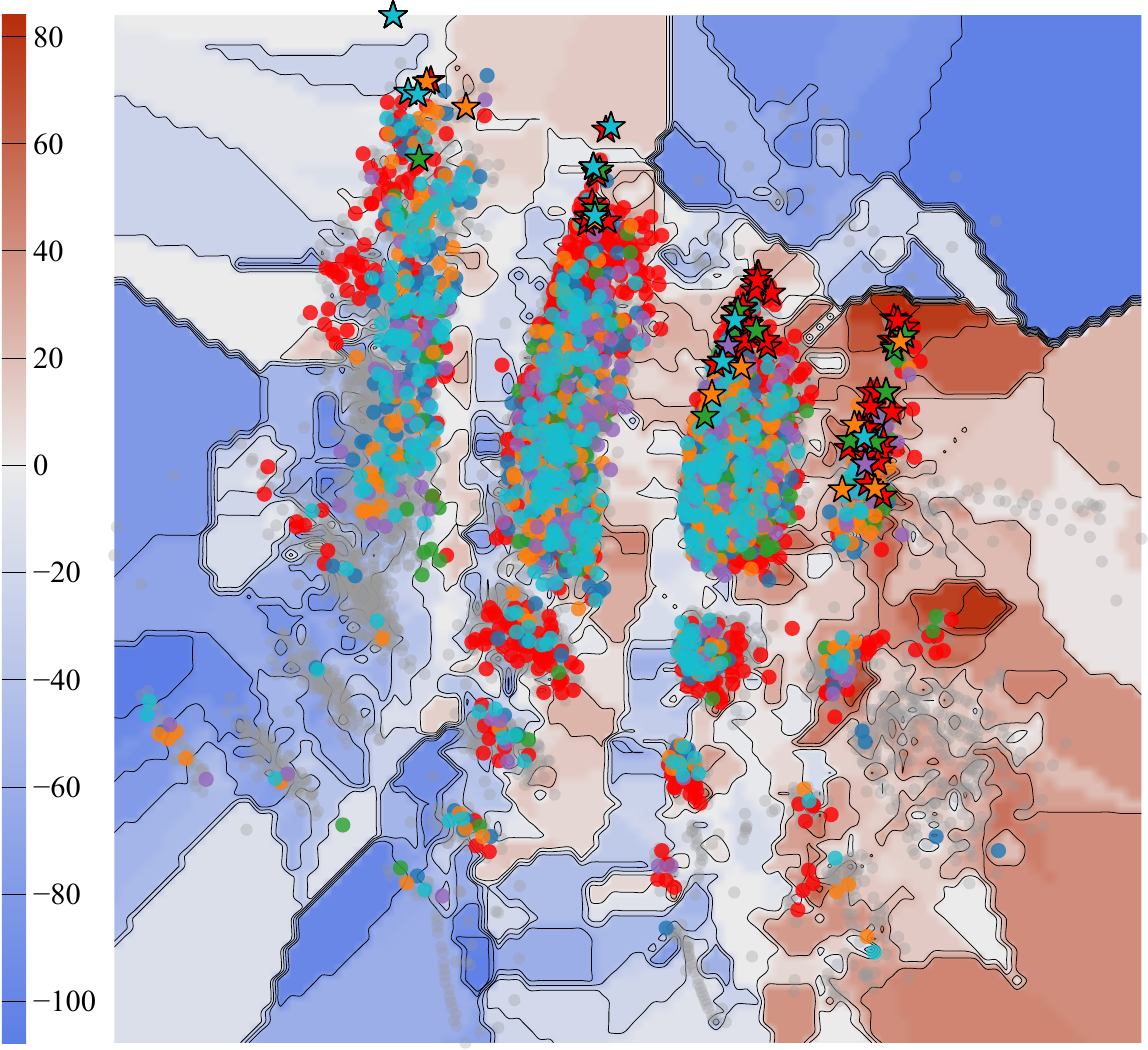}
    \small{(b) sce-2}
  \end{minipage}%
  \begin{minipage}{0.28\textwidth}
    \centering
    \includegraphics[width=\columnwidth]{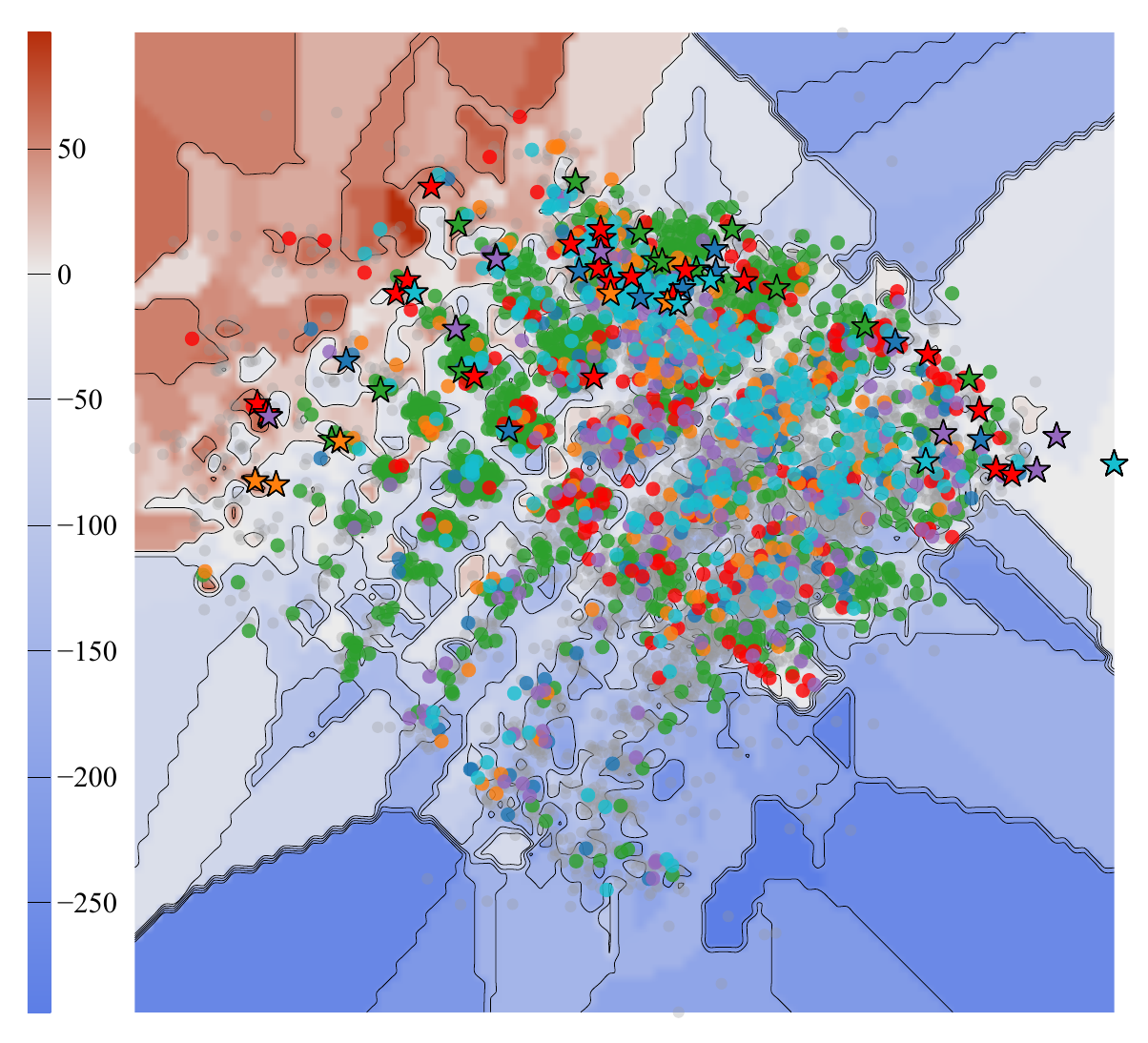}
    \small{(c) sce-3}
  \end{minipage}%

  \medskip

  \begin{minipage}{0.28\textwidth}
    \centering
    \includegraphics[width=\columnwidth]{MarineMicro_MvsM_4_mirror_state_kde_band_tight_single.pdf}
  \end{minipage}%
  \begin{minipage}{0.28\textwidth}
    \centering
    \includegraphics[width=\columnwidth]{MarineMicro_MvsM_4_dist_mirror_state_kde_band_tight_single.pdf}
  \end{minipage}%
  \begin{minipage}{0.28\textwidth}
    \centering
    \includegraphics[width=\columnwidth]{MarineMicro_MvsM_8_mirror_state_kde_band_tight_single.pdf}
  \end{minipage}%

  \begin{minipage}{0.28\textwidth}
    \centering
    \includegraphics[width=\columnwidth]{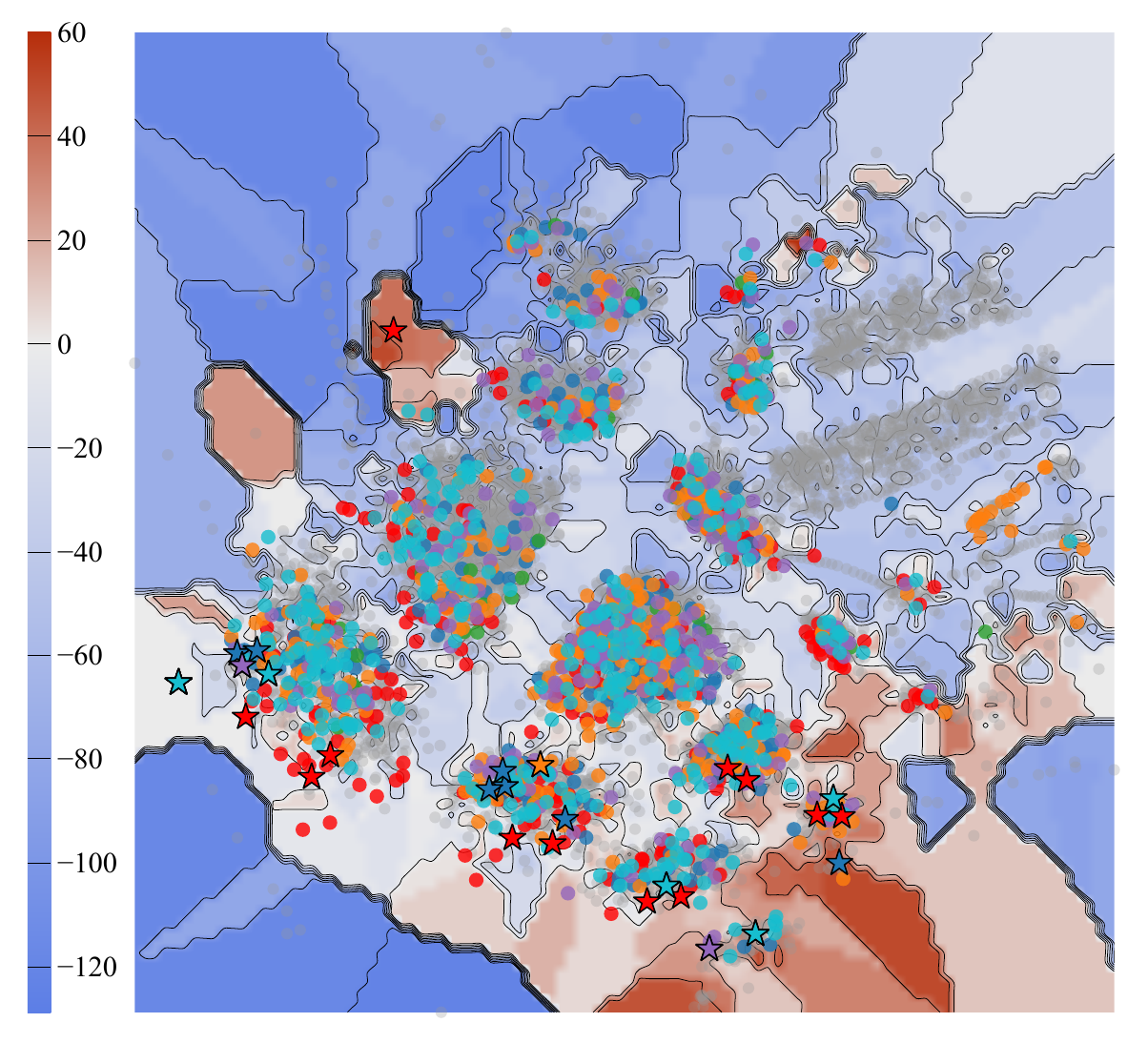}
    \small{(d) sce-1m}
  \end{minipage}%
  \begin{minipage}{0.28\textwidth}
    \centering
    \includegraphics[width=\columnwidth]{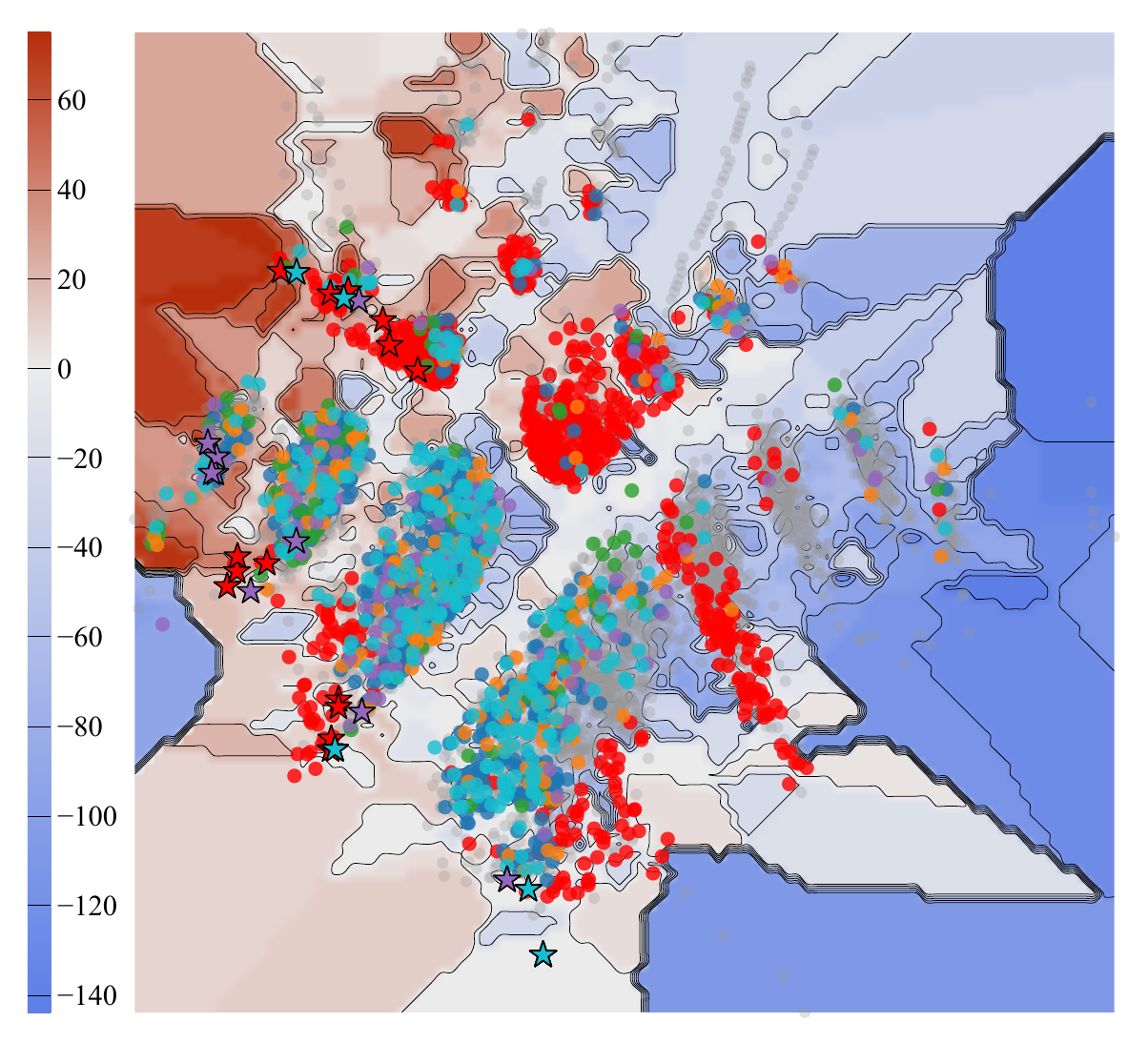}
    \small{(e) sce-2m}
  \end{minipage}%
  \begin{minipage}{0.28\textwidth}
    \centering
    \includegraphics[width=\columnwidth]{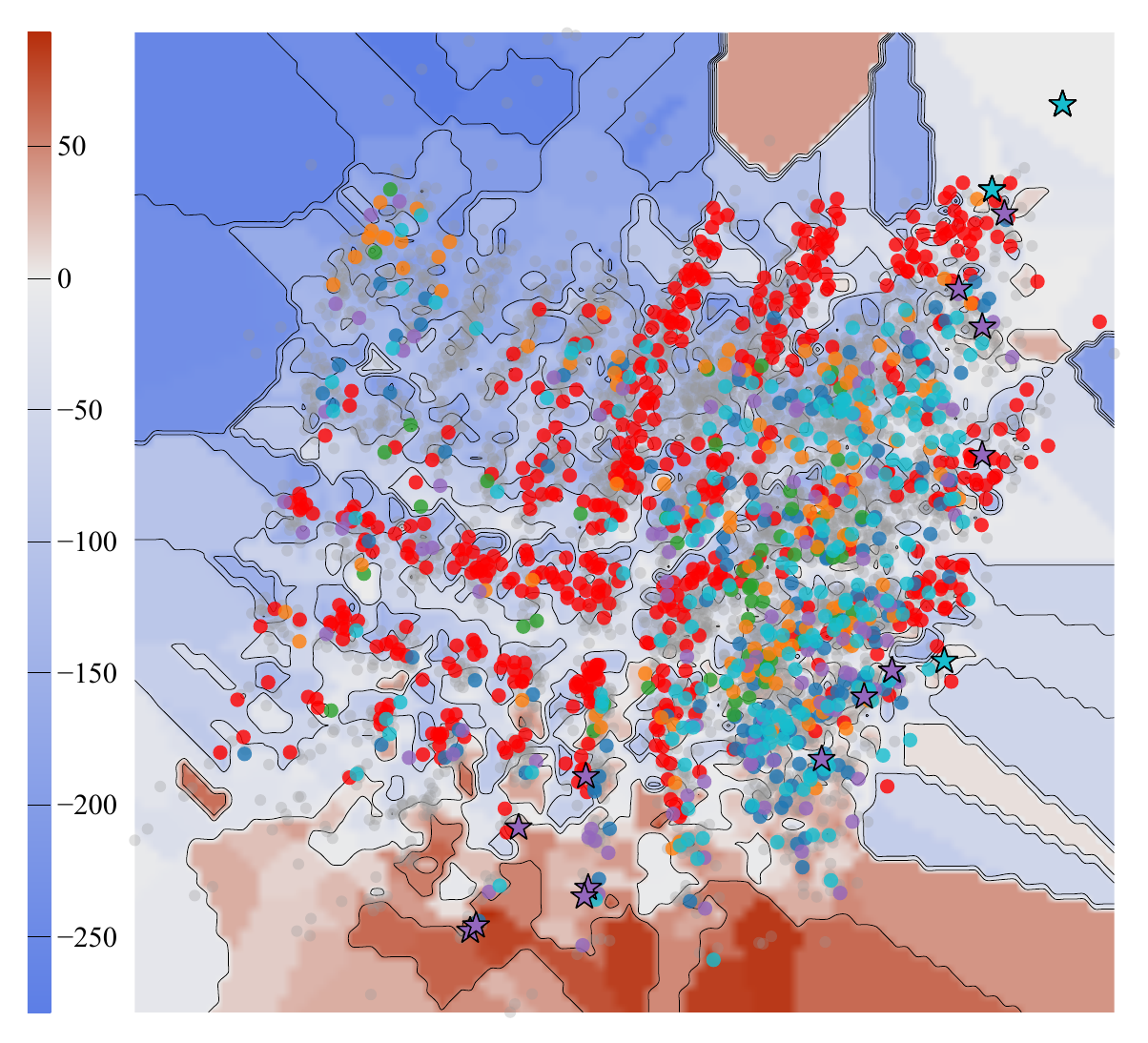}
    \small{(f) sce-3m}
  \end{minipage}%

  \caption{State-sampling frequency weighted density ($95.4\%$ HDI, $\approx\pm 2\sigma$) and distribution of sampled states over state value landscape for different algorithms across all scenarios.}
  \label{fig:state sampling frequency appendix}
\end{figure}

For state value landscape distribution, the state space is projected onto a two-dimensional plane via multidimensional scaling (MDS) and the mean state value is interpolated to yield a continuous landscape. The color range of contour maps transitions from blue (low-value, unfavorable states) to red (high-value, favorable states). Coloured dots denote states visited exclusively by a single algorithm, whereas gray dots indicate states visited by more than one algorithms. Black-edged stars highlight the states contained in the best samples of each algorithm.

Attributed to the influence map hashing, the distribution of states sampled by different algorithms can be readily visualized and analyzed. Viewed from an alternative perspective, the original state space depicted in Figure~\ref{fig:state sampling frequency appendix} is characteristically vast, rugged and complex. The influence map hashing achieves an effective compression of the state space by leveraging the battlefield situation, while the accuracy loss induced by this compression is offset by efficient exploration driven by cluster-based scripts. The proposed HRL-IM/CBS does not incur excessive visits to low-value regions as a consequence of its state-space processing. Instead, through the synergistic interaction between the co mponents, the proposed algorithm retains the ability to approach high-value regions.

It should be noted that the details of game state clustering, the distance metric between states, and the landscape visualization based on the state-distance matrix are still under investigation and are therefore not elaborated in this paper. Nevertheless, the tool suffices to illustrate the sample efficiency of the proposed method. From the figure, it is evident that, despite the rugged landscape, the states explored by HRL-IM/CBS are located closer to high-value regions, and the sampling density is skewed toward larger state values, demonstrating that the algorithm avoids low-reward regions and gravitates toward promising areas during training.

\end{document}